# Row-sensing Templates: A Generic 3D Sensor-based Approach to Robot Localization with Respect to Orchard Row Centerlines

Zhenghao Fei[1], Stavros Vougioukas[1]

[1]University of California, Davis, Department of Biological and Agricultural Engineering

**Abstract:**

Accurate robot localization relative to orchard row centerlines is essential for autonomous guidance where satellite signals are often obstructed by foliage. Existing sensor-based approaches rely on various *features* extracted from images and point clouds. However, any selected features are not available consistently, because the visual and geometrical characteristics of orchard rows change drastically when tree types, growth stages, canopy management practices, seasons, and weather conditions change. In this work, we introduce a novel localization method that doesn't rely on features; instead, it relies on the concept of a *row-sensing template*, which is the expected observation of a 3D sensor traveling in an orchard row, when the sensor is anywhere on the centerline and perfectly aligned with it. First, the template is built using a few measurements, provided that the sensor's true pose with respect to the centerline is available. Then, during navigation, the best pose estimate (and its confidence) is estimated by maximizing the match between the template and the sensed point cloud using particle-filtering. The method can adapt to various orchards and conditions by re-building the template. Experiments were performed in a vineyard, and in an orchard in different seasons. Results showed that the lateral mean absolute error (MAE) was less than 3.6% of the row width, and heading MAE was less than 1.72°. Localization was robust, as errors didn't increase when less than 75% of measurement points were missing. The results indicate that template-based localization can provide a generic approach for accurate and robust localization in real-world orchards.

# 1 Introduction

Labor cost and an increasing farm labor shortage are two main drivers for developing and deploying mechanization and automation technologies in orchards and vineyards (Zhang, 2017; Charlton et al., 2019). A third, significant driver is the need to implement precision horticulture, i.e., executing operations such as spraying, thinning, pruning, and harvesting while taking into consideration an orchard's or vineyard's spatial and temporal variability. Precision horticulture can increase the efficiency of resources, and consequently reduce cost and negative ecological impact (Zude-Sasse et al., 2016).

Agricultural robots can help ease orchard labor shortages by either replacing workers in labor-intensive tasks like harvesting (e.g., Williams et al., 2020) or assisting human workers in various orchard production activities (e.g., harvesting, pruning, spraying, and mowing) to increase working efficiency (Zhang, 2017). For example, the utilization of an autonomous utility vehicle resulted in efficiency gains of up to 58% (Bergerman et al., 2015) for tasks such as pruning conducted on the top part of trees when compared with the same task performed on ladders. Agricultural robots can also provide advanced sensing, computation, and actuation that facilitates precision horticulture. Example applications include selective spraying, where chemical inputs are reduced dramatically (Zhang, 2017; Asaei et al., 2019) and selective pruning (Botterill et al., 2017).

Robot operation in orchards relies on accurate localization inside the rows of orchard blocks, and at the end-of-block headland spaces (Figure 1a). Inside orchard rows, robots travel along the row centerlines, and therefore, auto-guidance requires knowledge of the robot's *position along the row's centerline*, and its *lateral displacement and heading offsets relative to this centerline*. In the headlands, robots execute appropriate turning maneuvers to move to another row or move to another orchard block.

Unlike localization in open fields where cm-level accurate GNSS (Global Navigation Satellite System) signals are available, accurate and robust localization inside orchard rows cannot rely solely – or at all - on GNSS. The reason is that the foliage of tall trees (Figure 1a, b) often blocks GNSS signals or introduces multipath effects, rendering satellite-based localization with a moving GNSS receiver impossible or unreliable. This effect may not be as severe in the headlands between orchard blocks, although it can be present. Therefore, localization methods must utilize sensors that take local measurements of the surrounding environment (trees, ground, sky, irrigation lines, etc.). Visual cameras, depth or RGB-D cameras, and LiDARs are sensors that are commonly used in orchards to collect reflected energy from the surrounding environment, with high spatial resolution.

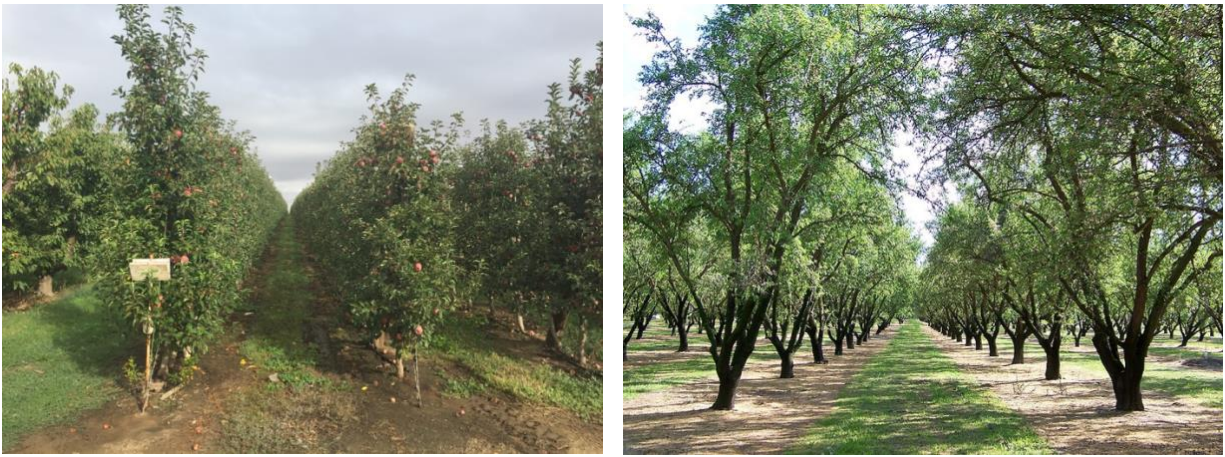

*Figure 1. a) Left: Rows of high-density trellised apple trees belonging to an orchard block; headland space is shown in the front (Lodi, California, 2018); b) Right: Rows of almond trees (Winton, California, 2015, courtesy of UC ANR).*

The rows of modern commercial orchard blocks are characterized by flat terrain and uniformity in tree types, placement/spacing and canopy shapes and sizes. The latter are the result – to a large extent - of the orchard owners' uniform canopy management practices. The orchard floor and the tree canopies on the left and right side of a row create a "tunnel"-shaped structure along the row. In densely planted fruit-wall type orchards, such as the one shown in Figure 1a, the cross-section of the "tunnel" is pretty much consistent along its centerline. In more sparsely planted orchards with non-planar canopies, such as the one in Figure 1b, the cross-section undulates in a repetitive fashion. The terms "consistent" and "repetitive" should not be interpreted in their strict mathematical sense of "equal" and "periodic", respectively, as factors such as biological variability in canopy shape and growth, imperfect hedging and pruning/trimming, missing trees or smaller/different trees (e.g., pollinators) introduce some variability, even within the same row. An important characteristic of the tunnel structures is that they do not change significantly among different rows or blocks in the same orchard, exactly because of the uniformity conditions presented in the beginning of the paragraph.

When a forward-looking 3D sensor is used for robot localization inside a row, and the sensor is placed near the ground level, below the treetops, the sensor will generate a 3D point cloud (also referred to as 'sensor readings') that captures a segment of the tunnel structure formed by the ground and the tree canopies. Depending on the sensor, the captured tunnel segment will extend over some distance (e.g., 10 to 20 m) in the front of the robot. These sensor readings exhibit two main characteristics.

The first characteristic is that the spatial distribution of the sensor readings does not change significantly when the sensor's reference frame translates along/parallel to the orchard centerline. The reason lies in the consistent or repetitive nature of the tunnel's shape. A major consequence of this characteristic is that absolute localization along the centerline - with respect to a reference entry point of the row - does not seem feasible/practical using local, near-ground sensing. Simply put, if a 3D sensor were placed one quarter or halfway inside a row, it could not tell the difference using local measurements. For this reason, in the absence of reliable GNSS signals or artificial landmarks that could be used for absolute positioning, existing literature (Section 2) shows that researchers have used various forms of odometry (e.g., wheel, inertial, or vision-based) for localization along the centerline from a reference starting point on the centerline (see Shalev & Degani, (2020) for a different approach). It should be pointed out that this characteristic becomes weaker as the robot approaches the end of the row, because the sensor captures readings from the space beyond the end of the row, which typically looks very different than the tunnel space inside the row (e.g., it contains headland space that is empty of trees, another orchard block with different trees or tree pattern and orientation, etc.).

The second characteristic is that sensor readings depend on the sensor frame's rotation or lateral translation (offset) with respect to the centerline of the orchard row. This characteristic suggests that lateral and rotational localization with respect to the centerline are possible using only local sensing. In fact, the published methods for such localization rely on a stronger assumption on this characteristic, i.e., that selected features of the orchard environment are distributed *symmetrically* about the centerline, and that these features can be detected and localized reliably (see detailed literature review in section 2). These features are used to estimate the robot's lateral displacement and heading offsets relative to the centerline. Proposed features include the tree trunks on the left and right sides of the row, the intersections between trunks and ground, the sky region, the orchard floor, or the planes of flat, fruiting-wall type tree canopies.

The main shortcoming of feature-based methods is that they have not been shown to generalize well or be robust enough in different orchard settings. Orchards constitute diverse, complex, and dynamic environments. Tree type, age, placement, and architecture, as well as canopy and orchard floor management practices can severely affect the presence, appearance, and symmetry (about the centerline) of features. Seasonal variation (e.g., dormant vs. blooming trees), weather, and illumination conditions, missing trees also introduce variability in feature appearance, symmetry, or even availability. To the authors' knowledge, a general approach that does not rely on specific features, or centerline symmetry assumptions, and can adapt easily to a large range of orchard environments is not available.

In this paper, we propose a localization method which adopts a probabilistic framework, uses raw 3D point clouds, instead of features extracted from the raw data, and is shown to be generic and robust. The proposed method consists of two stages.

In its first stage, the method capitalizes on characteristic #1 and builds a uniform 3D grid that stores in each voxel the probability that this voxel is occupied, when the sensor is anywhere on the row's centerline and aligned to it. Essentially, the 3D grid – referred to as a "row-sensing template" or "template" – represents what the sensor expects to "sense" if it is placed at any point on the centerline, with its frame aligned with it; it is an occupancy grid (Elfes, 1989) for the space inside the sensor's field of view (not the world). The template is built using a small set of point cloud measurements. The major requirement during the template-building stage is that the sensor's lateral and heading offsets (ground truth) are known.

In its second stage, the method capitalizes on characteristic #2 and uses the template for Monte-Carlo localization in real-time. A static measurement model is implemented that returns the probability of the current point-cloud measurement given the template and a proposed pose. The vehicle pose is estimated as the pose that maximizes the probability of getting the observed measurement. The space of possible poses is searched by generating uniformly distributed random poses or by integrating the measurement model with a particle filter framework.

The main contributions of this paper toward this goal are the following:

1) It introduces the concept of the 'row-sensing template' and utilizes it to develop a new generic and robust sensor-based method to estimate a vehicle's heading and lateral offset with respect to the row centerline in orchards; the method capitalizes on characteristics #1 and #2, but does not rely on features and symmetry assumptions.

2) It presents extensive experimental results in two different orchards and various seasons to evaluate the proposed method's localization accuracy and robustness, analyzes the method's performance as the robot approaches the end of the row, where characteristic #1 is violated, to increasing extent, i.e., sensor readings start varying under translation along the centerline. The paper also presents the comparison to two of the existing pointcloud based in-row localization methods.

The rest of this paper is organized as follows: In section 2, we present the related works, and in section 3, we discuss in detail our proposed approach. Next, in section 4, we present the experimental platform and methods used for the experimental evaluation of the method. The results from our experiments are presented and discussed in section 5, and in section 6, we summarize our conclusions and discuss future work.

## 2 Related Work

Sensor-based localization inside orchard rows has been addressed by many researchers, with cameras (monocular and stereo) and LiDARs (2D and 3D) being the most commonly used sensors. The main idea behind the existing methods is to utilize specific visual or geometrical features or structures to estimate directly or indirectly the row's centerline and localize the sensor (robot) with respect to it. Barawid et al. (2007) used a 2D LiDAR scanner to detect tree trunks and the Hough-transform to extract the left and right tree lines independently; then, he computed the corresponding centerline to determine the pose of the vehicle. Similarly, He et al., (2010) proposed a machine vision algorithm to detect tree trunks and *the boundaries between the trunks and the ground, to* estimate tree row lines, and the corresponding row centerline. Hamner et al., (2011) used a 2D LiDAR to detect tree trunks and the Hough transform to detect right and left lines that are constrained to be parallel; the centerline was computed from them. Marden et al., (2014) estimated grapevine trunk lines using the RANSAC method and used these lines as features in a line-based EKF-SLAM framework; their method can simultaneously do localization and mapping (line map). Bell et al., (2016) used a 3D laser scanner to measure the positions of posts and trunks in pergola-structured orchards (e.g., for kiwis) and calculate the row direction and centerline. Lyu et al., (2018) also proposed a method to detect the boundaries between trunks and the ground and used a naive Bayesian classifier for the free space centerline detection. Durand-Petiteville et al., (2018) presented a stereo vision-based method to find tree trunks by detecting their "shadows," i.e., concavities in the range component of the obtained point cloud. Other researchers have used the *ground*, *sky*, or *tree foliage* as features, and segmented them in image space to estimate the row's centerline. Subramanian et al., (2006) proposed to use RGB thresholding to segment the tree canopy in the image and find boundary lines; their method also combined a 2D lidar to detect a path using distance thresholding to increase their system robustness. Torres-Sospedra et al., (2011) used a multi-layer feedforward neural network to segment land/soil, sky, tree crown, and trunk areas in an image and then applied a Hough transformation on the borders between land and trees to determine the centerline of the path. Sharifi et al., (2015) improved the segmentation method by using the mean-shift algorithm to do clustering, along with a novel classification technique based on graph partitioning theory to classify clusters. Radcliffe, Cox & Bulanon (2018) used an upward looking camera to detect sky and tree canopy features for localization.

Zhang et al., (2013) proposed a method that utilizes 3D point clouds for localization. They divided the 3D point cloud into a left and a right set. Then, they randomly selected points in both sets to compute multiple pairs of parallel-line features and used RANSAC to get the pair with the smallest number of outliers. The heading was directly computed from the best pair of the parallel lines. However, the method needed an additional step to segment tree trunks and large branches from the point cloud to accurately determine the lateral offset, because trunks and branches generate denser and more stable LiDAR returns than leaves and grass.

Stefas et al., (2016) proposed a vision-based UAV navigation method in orchards. Two approaches were proposed for localization. One approach used a monocular camera to detect strips of clean soil along the tree trunk lines, next to mowed grass inside the rows, in what the authors referred to as "marked or clearly visible" tree rows or "road lines" in "well-maintained" orchards; straight parallel lines were then fit to the tree row pixels and the centerline was computed from them. The second approach used a binocular camera and the resulting pointcloud to estimate the ground plane and the UAV's attitude. Next, the tree lines were detected using the algorithm and assumptions of the monocular-camera approach. This approach relies heavily on the assumption that the orchard floor is maintained in such a good condition that the tree line is easily detectable as a strip of soil - clean from weeds or grass – against the green mowed floor inside the row. Stefas et al., (2019) extended their previous work to include branch detection for reactive UAV navigation inside orchard rows; the same approach was used to estimate the UAV's pose relative to the centerline.

All the above methods are applicable when the corresponding features they rely on are visible/detectable. As it was explained in Section 1, this may not be true in many cases. For example, tree trunks or trunk-ground intersections may not be visible; the sky may not be visible, or the orchard floor may be covered (Figure 2a, 2b, 2c). Blok et al., (2019) used a 2D LiDAR and a particle filtering approach for localization in orchard rows, without relying on features; however, it was assumed that tree trunks were the only objects scanned by the LiDAR. The method was found to be accurate and robust when some trees were missing. However, their probabilistic 2D LiDAR sensor model relied heavily on an accurate *a-priori* environmental model of the orchard structure (tree inter-row and intra-row spacing, tree spacing and trunk size) at each side of the robot, which would have to be re-developed for different tree architectures and orchard spacings. Furthermore, approaches that use a 2D LiDAR as the main sensor can only get single-plane information in space. Because tree canopies are three dimensional and irregular, and ground can be uneven and have grass, methods using 2D LiDAR are not as robust.

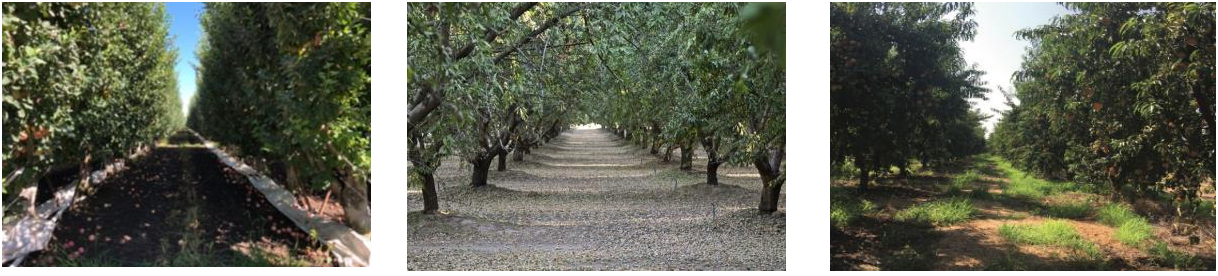

*Figure 2. Left: Apple tree trunks are hidden by foliage, and trunk-ground intersections are partially occluded by reflective tarps (Lodi, California, 2019); Center: The sky is not visible in an almond orchard (California, 2017); Right: The orchard floor is covered with patches of grass (Vougioukas, 2019).*

As a conclusion, accurate and robust methods for localization with respect to orchard row centerlines are still needed. Such methods should not depend on over-simplifying, extensive or unrealistic assumptions about orchard structure or the presence of features and should be applicable and easily adaptable in different types of orchards, and different seasons.

# 3 Template-based Localization

The proposed template-based localization method consists of two main stages. During the first template-building stage, a 3D sensor template **T** is constructed that represents what a 3D sensor would *expect to measure* (voxel occupancy frequency) if it was placed at various positions along the row's centerline with zero lateral and heading offset. During this phase, point cloud measurements with corresponding known poses (ground truth) are used to build the template. In the second phase, as new sensor readings are collected in real-time, a novel template-based sensor measurement model is used to compute the pose that results in the best alignment between the template and the translated and rotated readings.

The reasoning behind the template method is based on the following. Let a sensor travel (on a robot) inside a specific row $R$. The sensor has traversed distance $x$ from a reference point at the beginning of the row. The sensor pose with respect to the centerline of this row is $(x, y, \theta)_R$, where $y, \theta$ are the lateral displacement and heading offsets relative to the centerline. From characteristic #2 (see Introduction) we know that the sensor readings $\mathbf{S}(x, y, \theta)_R$ depend on the sensor frame's rotation and lateral translation (offset) with respect to the centerline of the row. If at any specific but unknown sensor pose $(x, y, \theta)_R$ one could record the point cloud at pose $\mathbf{S}(x, 0, 0)_R$, then localization relative to the centerline $(y, \theta)_R$ inside row $R$ could be estimated using some point cloud alignment method between $\mathbf{S}(x, y, \theta)_R$ and $\mathbf{S}(x, 0, 0)_R$. Clearly, this is not possible; however, characteristic #1 says that the spatial distribution of the sensor readings does not change significantly when the sensor translates along the row (change in $x$). Furthermore, the spatial distribution of the sensor readings will not change significantly in different rows (change in $R$). Based on these two characteristics a template can be thought of as an expected sensor measurement $\mathbf{T}(\{x\}, 0, 0)_{\{R\}}$ generated off-line at some set of one or more rows $\{R\}$ and a set of positions $\{x\}$, which can be used to estimate on-line $(y, \theta)_R$ for any row.

Next, basic terms, symbols, and coordinate frames are introduced that will be used extensively in the rest of the paper. Let $\{R\}$ be the frame of the currently traversed orchard row. Its origin lies on the centerline of the row, between the first pair of trees at the beginning of the row (Figure 3). Its x-axis is the centerline of the row and points toward the other end of the row (forward), and its z-axis is perpendicular to the ground and points upward. $\{V\}$ is the vehicle frame, with its x-axis pointing forward, and the z-axis pointing upward. Let $\{C\}$ be the 3D sensor frame. In this work, for simplicity, $\{C\}$ coincides with $\{V\}$, and both may be used interchangeably. (In general, $\{C\}$ and $\{V\}$ are connected via a known rigid body transformation.)

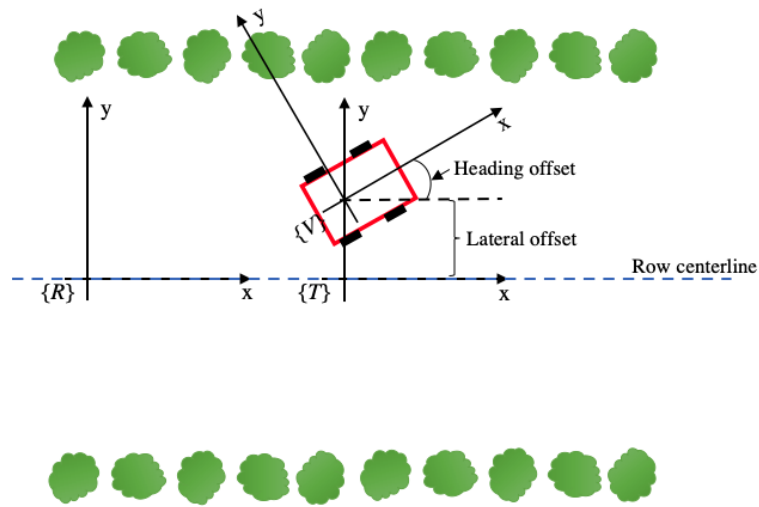

*Figure 3 The figure shows the orchard row frame {R}, the vehicle frame {V}, the template frame {T}, and the definition of lateral and heading offset.*

An orchard row template **T** represents the expected 3D sensor measurement, when the sensor is on the centerline, and aligned to the row frame $\{R\}$'s x axis. Therefore, the row template frame $\{T\}$ – by definition - has the same orientation as $\{R\}$, and its origin lies on the centerline, and at the same x coordinate as $\{V\}$ with respect to $\{R\}$.

The template is represented as a uniform 3D grid made up of voxels defined over a continuous space in a template frame $\{T\}$; each voxel has a value that represents the occupancy frequency of that voxel. Each measurement is a point cloud triangulated from a pair of stereo images. At time $t$, let the n$^{th}$ point of the point cloud be denoted as $^{\{V\}}z_t^n$, and the set of all points, i.e., the point cloud itself be $^{\{V\}}Z_t$. Our goal is to estimate the vehicle's pose with respect to the row frame $^{\{R\}}X_t = [\,^{\{R\}}x_t,\,^{\{R\}}y_t,\,^{\{R\}}z_t,\,^{\{R\}}\alpha_t,\,^{\{R\}}\beta_t,\,^{\{R\}}\theta_t]$, where $\alpha$, $\beta$, and $\theta$ are roll, pitch, and yaw, respectively. The distance along the centerline $^{\{R\}}x_t$ can be obtained by odometry and hence is outside the scope of this work. The quantities $^{\{R\}}z_t$, $^{\{R\}}\alpha_t$, $^{\{R\}}\beta_t$ are estimated by finding the ground plane as in section 3.1. The main focus of this paper is to estimate $^{\{R\}}y_t$ and $^{\{R\}}\theta_t$, using our template-based method. Next, the two stages of our method are presented in detail.

Stage 1: Given a set of point cloud measurements $^{\{V\}}Z_{t_1,t_2,\cdots t_n}$ with corresponding known lateral offsets $^{\{R\}}y_{t_1,t_2,\cdots t_n}$ and headings $^{\{R\}}\theta_{t_1,t_2,\cdots t_n}$ with respect to the centerline, build an row-sensing template **T**.

Stage 2: Given a sequence of point cloud measurements $^{\{V\}}Z_{t_1,t_2,\cdots t_n}$, visual odometry information $^{\{R\}}u_{t_1,t_2,\cdots,t_n}$, and the pre-built orchard row template **T**, compute the vehicle's lateral offset $^{\{R\}}y_t$ and heading $^{\{R\}}\theta_t$ with respect to the row's centerline, at each time step. The computational pipelines of both stages are shown in Figure 4. Each individual module is explained in detail in this section.

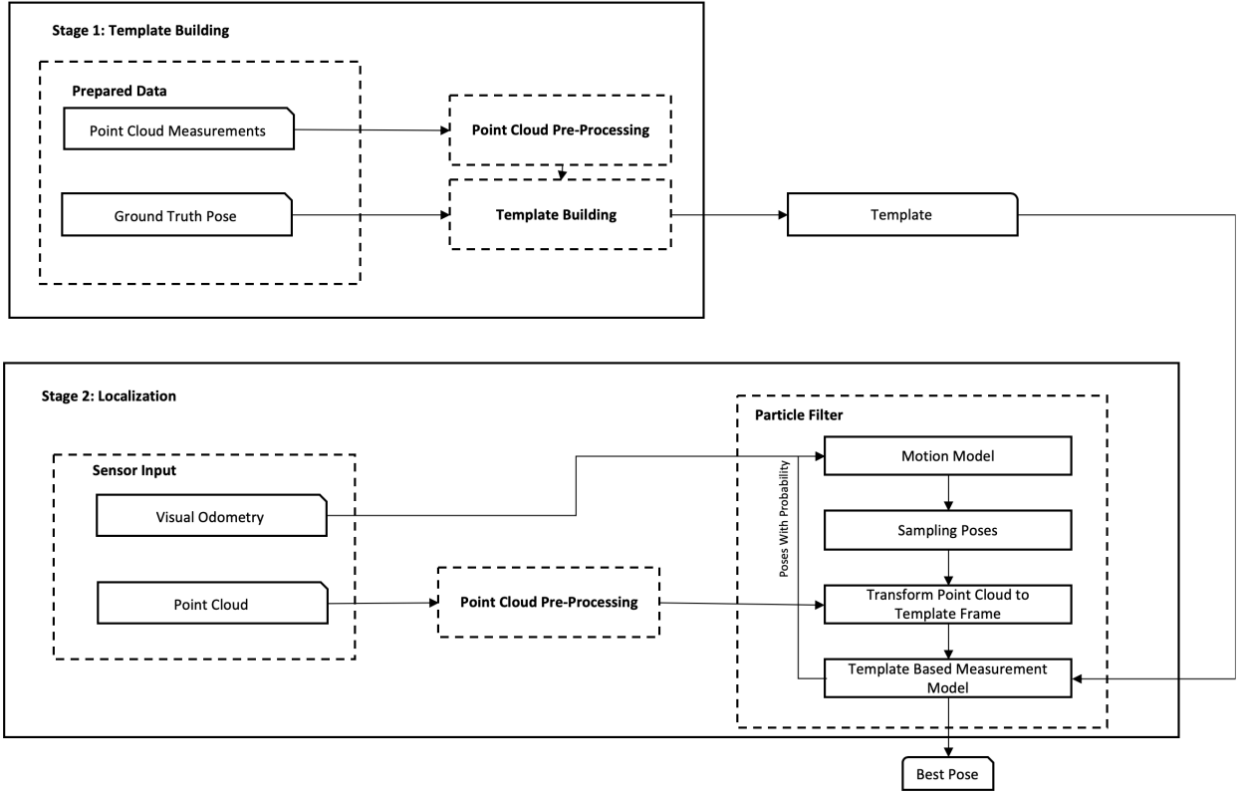

*Figure 4 The figure shows two stages of the template-based localization approach ("Template Building" and "Localization") and their corresponding computational pipelines.*

## 3.1 Stage 1: Template Building

### 3.1.1 Template building with ground truth pose

The template **T** represents in a 3D grid the sensor measurement *expectation* (voxel occupancy frequency) if it were placed on the row's centerline with zero lateral and heading errors. Here we propose a way to build an orchard row template – starting from an empty template - using a set of measurements $^{\{C\}}Z_{t_1, t_2, \cdots t_n}$ with corresponding known lateral offsets $^{\{R\}}y_{t_1, t_2, \cdots t_n}$ and headings $^{\{R\}}\theta_{t_1, t_2, \cdots t_n}$.

As a first step, each point cloud measurement $^{\{C\}}Z_t$ is input to the Point Cloud Pre-Processing module (Figure 5). In this module, the point cloud is first down-sampled using a voxel down-sampling method (Rusu et al., 2011). All points belonging to a voxel are represented by a single point – their centroid, and the point cloud is transferred into $\{V\}$. The assumption is made that the orchard ground is relatively flat, and the ground plane can be extracted in the point cloud measurement, so that the vehicle's roll and pitch can be easily recovered. So, RANSAC (Bolles et al., 1981) is used to estimate the ground plane and the vehicle's roll $\alpha_t$, pitch $\beta_t$, and $z_t$ states in $\{R\}$.

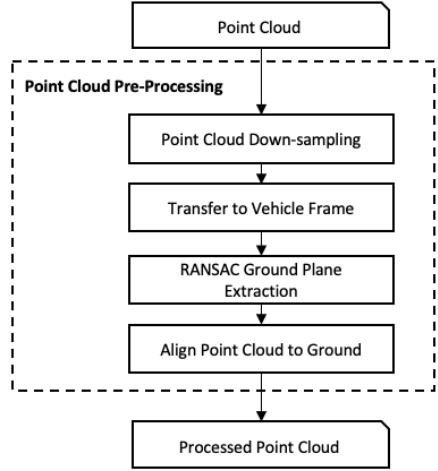

Figure 5 The figure shows the point cloud pre-processing pipeline.

As a second step, using the states $^{\{R\}}\alpha_t$, $^{\{R\}}\beta_t$, $^{\{R\}}z_t$ estimated by RANSAC and the known states $^{\{R\}}y_t$, $^{\{R\}}\theta_t$, the measurements $^{\{C\}}Z_t$ is transformed to the template frame $\{T\}$. All the points that do not belong to the current tree row and are outside the sensor's range are discarded by applying a spatial rectangular cutoff filter (**CutoffFilter**()) of appropriate dimensions set by the row width, the maximum tree height and the sensor's maximum range (see section 5.1 for the values used in this work).

The third step (implemented in **UpdateTemplate**()) updates the contents of the template's voxels using the point cloud $^{\{T\}}Z_t$ by incrementing the current value of each voxel whose volume contains a point from $^{\{T\}}Z_t$. Then, the occupancy *frequency* of each voxel is computed as an estimate of the voxel's occupancy probability. The region out of the row range is undefined, there might be another row next to the current or not, so we fill all voxels within the template - but not in the current row- with a value that corresponds to "no information on frequency" (*noInfoFrequency*). The entire process is given in **Algorithm 1**, and the related frame transformations are shown in Figure 6. Figure 7 shows the visualization of a row template.

**Algorithm 1 BuildTemplate**

**Input**: $^{\{C\}}Z_{t_1,t_2,...,t_n}$, $^{\{R\}}[y,\theta]_{t_1,t_2,...,t_n}$, templateResolution, templateRange, rowRange, noInfoFrequency
**Output**: template **T**
Create an empty template **T** (all zero 3D grid) with templateResolution and templateRange
$^{\{V\}}Z_{t_1,t_2,...,t_n}$, $^{\{R\}}[\alpha,\beta,z]_{t_1,t_2,...,t_n} = \textbf{PointCloudPreProcess}\left(^{\{C\}}Z_{t_1,t_2,...,t_n}\right)$
**for** t = 1 : n **do**
   $R_t = \textbf{RotationMatrixFromEuler}\left(^{\{R\}}\alpha_t, \, ^{\{R\}}\beta_t, \, ^{\{R\}}\theta_t\right)$
   $t_t = [0, \, ^{\{R\}}y_t, \, ^{\{R\}}z_t]$
   $^{\{V\}}T_{\{T\}} \in SE(3) = [R_t | t_t]$
   $^{\{T\}}Z_t = \, ^{\{T\}}T_{\{V\}} \, ^{\{V\}}Z_t = \, ^{V}T_T^{-1} \, ^{\{V\}}Z_t$
   $^{\{T\}}Z_t = \textbf{CutoffFilter}(^{\{T\}}Z_t, \textbf{\textit{rowRange}})$
   $\textbf{T} = \textbf{UpdateTemplate}(\textbf{T}, \, ^{\{T\}}Z_t)$
**end for**
$\textbf{T} = \frac{\textbf{T}}{n}$ # convert count map to frequency map
Fill the region out of rowRange in **T** with noInofoFrequency
**return T**

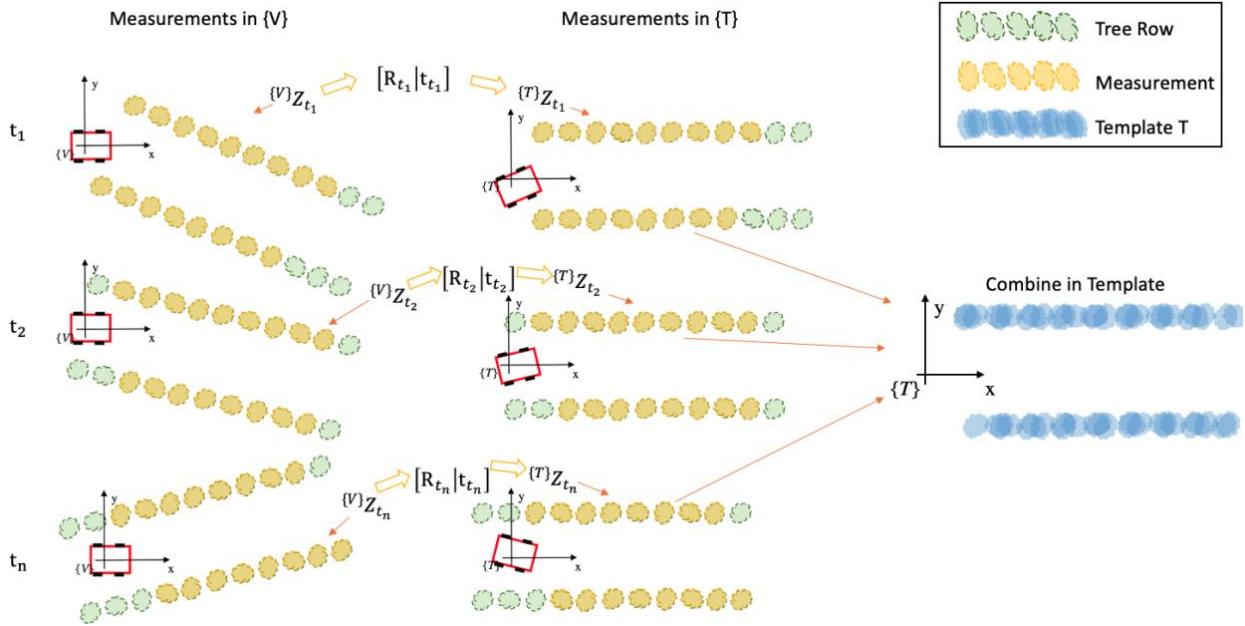

*Figure 6 Measurement frame transformations in Algorithm 1. Measurements are received in {V} by camera, they are transferred into {T} using the corresponding ground truth pose, then they are combined in a 3D template.*

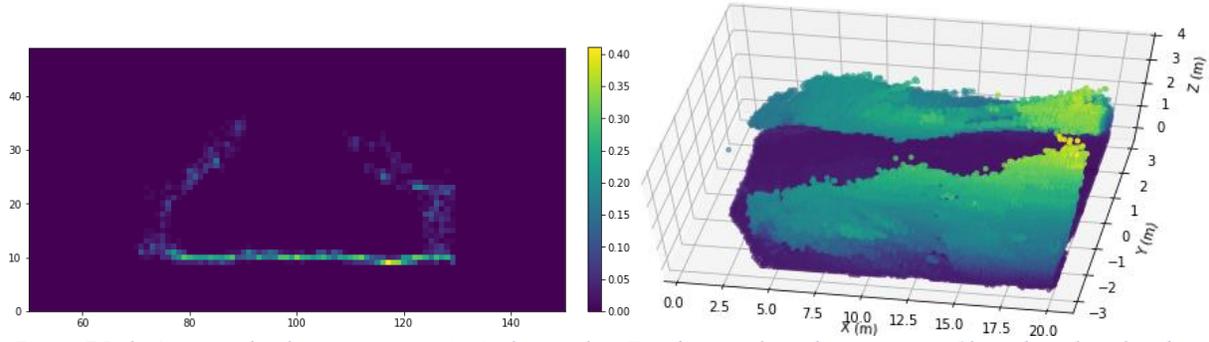

*Figure 7 Left: An example of a cross-section (y-z) of a template T at distance from the sensor, x = 12 m; the color of each voxel represents the occupancy frequency; Right: 3D visualization of the template with voxel occupancy frequency larger than 0.02.*

## 3.2 Stage 2: Localization using a template

### 3.2.1 Measurement Model

The measurement model (or sensor model) is the probability $P(^{\{V\}}Z | X, \mathbf{T})$ of getting a point cloud measurement $^{\{V\}}Z$, given the vehicle pose $X$ and the template $\mathbf{T}$ (Thrun et al., 2005). This probability can be obtained as the product of probabilities of all individual points under the assumption that individual point measurements are independent, given the vehicle pose $X$ and the template $\mathbf{T}$.

$$P(^{\{V\}}Z | X, \mathbf{T}) = \Pi_{k=1}^{n} P(^{\{V\}}z^k | X, \mathbf{T}) \quad (1)$$

The template $\mathbf{T}$ is built in a way that each voxel is an estimate of the probability of the sensor to get a measurement in that voxel in frame $\{T\}$. So, $\mathbf{T}$ is used as a likelihood field that can be indexed to get the individual measurement probability $P(^{\{T\}}z^k | \mathbf{T})$. Each individual measurement probability $P(^{\{V\}}z | X, \mathbf{T})$ can be calculated by transforming $^{\{V\}}z$ to $^{\{T\}}z$ and performing a table lookup for the probability in $\mathbf{T}$. The algorithm of this measurement model is shown in **Algorithm 2**, and its measurement model process is shown in Figure 8. Figure 9 shows $P(^{\{V\}}z^k | X, \mathbf{T})$ for each point in the measurement given a good pose proposal $X$ and a bad pose proposal; the overall point measurement probability is higher (brighter) for a good pose proposal.

---

**Algorithm 2 MeasurementModel**

**Input**: $^{\{C\}}Z_t$, $\mathbf{T}$, $^{\{R\}}X$, $rowRange$

**Output**: $P({}^{\{V\}}Z \mid {}^{\{R\}}X, \mathbf{T})$
${}^{\{V\}}Z_t, {}^{\{R\}}[\alpha, \beta, z]_t = \mathbf{PointCloudPreProcess}({}^{\{C\}}Z_t)$
$R_t = \mathbf{RotationMatrixFromEuler}({}^{\{R\}}\alpha_t, {}^{\{R\}}\beta_t, {}^{\{R\}}\theta_t)$
$t_t = [0, {}^{\{R\}}y_t, {}^{\{R\}}z_t]$
${}^{\{V\}}T_{\{T\}} \in SE(3) = [R_t | t_t]$
${}^{\{T\}}Z_t = {}^{\{T\}}T_{\{V\}} {}^{\{V\}}Z_t = {}^{V}T_T^{-1} {}^{\{V\}}Z_t$
${}^{\{T\}}Z_t = \mathbf{CutoffFilter}({}^{\{T\}}Z_t, \text{rowRange})$
$q = 1$
**for** k = 1 : n **do**
    p = retrieve probability of the point ${}^{\{T\}}Z_t^k$ from $\mathbf{T}$
    q = q * p
**end for**
$P({}^{\{V\}}Z \mid {}^{\{R\}}X, \mathbf{T}) = q$
**return** $P({}^{\{V\}}Z \mid {}^{\{R\}}X, \mathbf{T})$

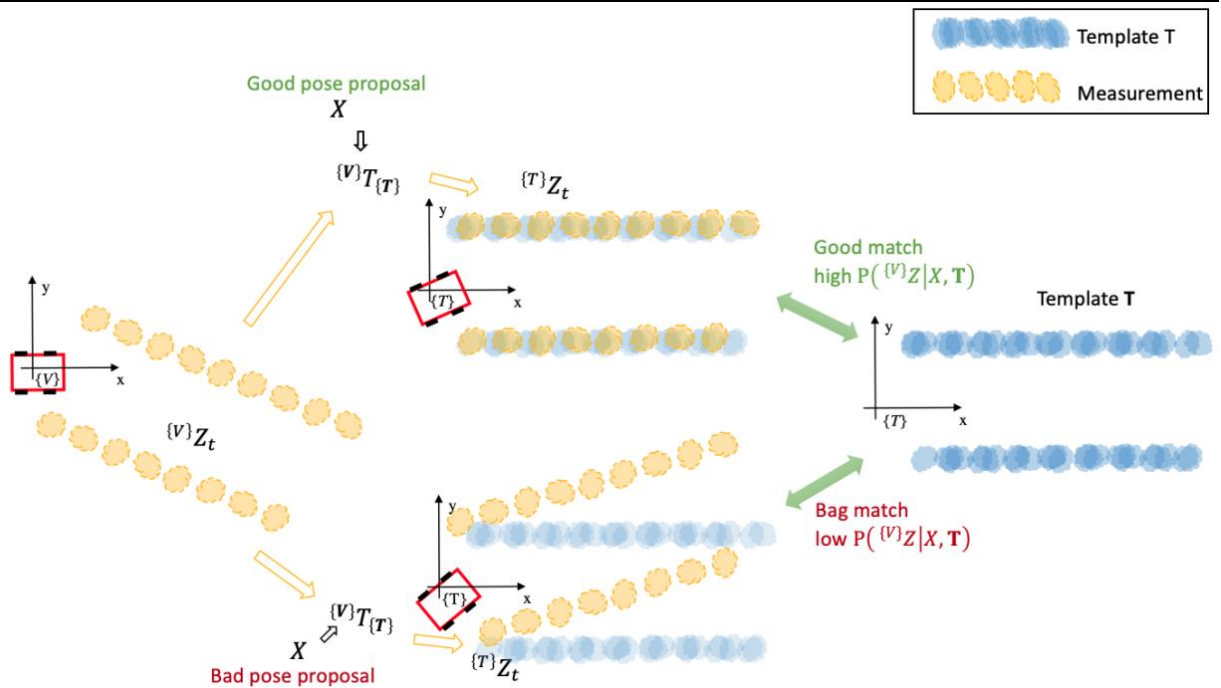

*Figure 8 This figure represents the measurement model process of the Algorithm 2. For a measurement ${}^{\{V\}}Z$, a good pose proposal can transform the measurement into {T} that well match the template T, thus resulting a high probability.*

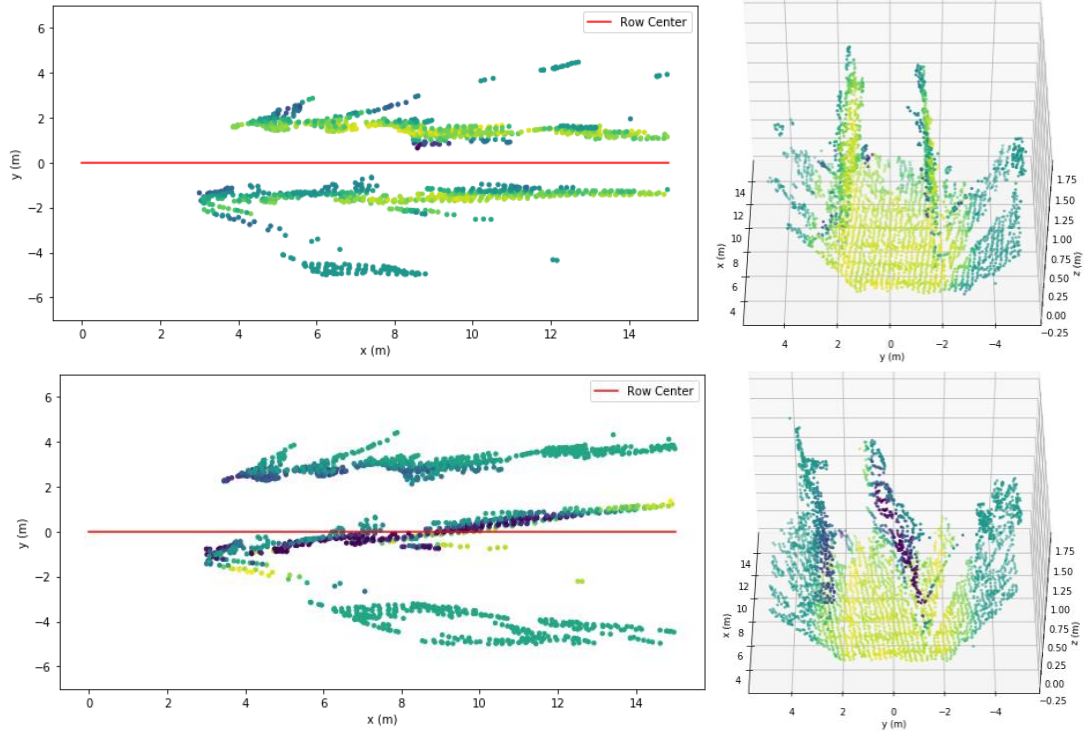

*Figure 9 Top: Top-down view (left) and 3D view (right) of $P\left({}^{\{V\}}z^k|X,T\right)$ for all the points in ${}^{\{V\}}Z$ of a good pose proposal. Bottom: Top-down view (left) and 3D view (right) of $P\left({}^{\{V\}}z^k|X,T\right)$ for all the points in ${}^{\{V\}}Z$ of a bad pose proposal. The brighter a point, the more likely this measurement point is correct.*

### 3.2.2 Monte Carlo Localization

Given the measurement model $P\left({}^{\{V\}}Z|X,\mathbf{T}\right)$, a template $\mathbf{T}$ and a measurement ${}^{\{V\}}Z_t$, the Monte Carlo (aka Particle Filter) Localization framework is used to estimate vehicle pose $X_t$ (Thrun et al., 2005). Under this framework, $n$ multiple possible poses $X_t^{[i]}$ are sampled from a distribution $\mathbf{D}$, to generate the set of sampled poses $\mathbf{X}_t = \left[X_t^{[1]}, X_t^{[2]} \ldots, X_t^{[n]}\right]$. The likelihood of each possible pose in this set is evaluated, and the pose with maximum likelihood is selected.

$$\hat{X}_t = \underset{X \in \mathbf{X}_t}{arg\ max} P\left({}^{\{V\}}Z_t|X,\mathbf{T}\right) \qquad (2)$$

An obvious choice for $\mathbf{D}$ is a uniform distribution in frame $\{T\}$ (e.g., $y \in U[-0.8m, 0.8m]$, $\theta \in U[-0.6rad, 0.6rad]$). The Monte Carlo Localization algorithm with sampling from the uniform distribution is given next, in **Algorithm 3**.

---

**Algorithm 3 MonteCarloLocalization**(Uniform sampling)

---

**Input**: ${}^{\{C\}}Z_t$ , **Uniform distribution U**, **T**, rowRange,

**Output**: $\hat{X}_t$, Pose covariance $\Omega_t$
**for** k = 1 : n **do**
   $X_t^k = \textbf{SamplePose}(U)$
   $w_t^k = \textbf{MeasurementModel}(^{\{C\}}Z_t, \textbf{T}, X_t^k, \text{rowRange})$
**end for**
$i = \underset{0 \leq k \leq 1}{\arg\max} \, w_t^k$
$\hat{X}_t = X_t^i$
$\Omega_t = \textbf{CalculateCovarianceMatrix}(X_t)$
**return** $\hat{X}_t, \Omega_t$

Figure 10 shows an example of the likelihood field of $X \in \textbf{D}$, where **D** is a uniform distribution.

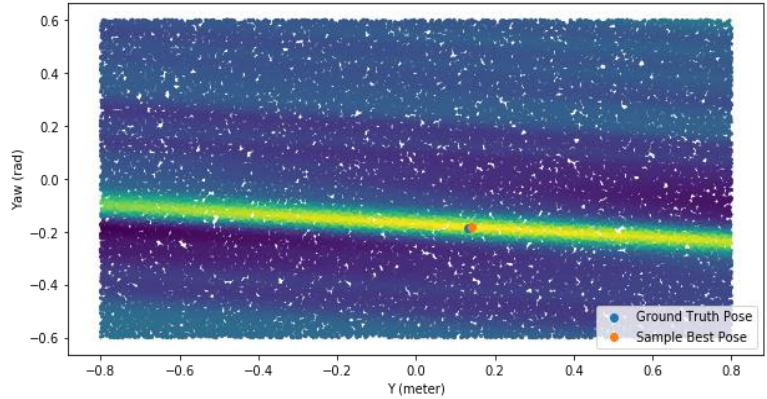

*Figure 10 The figure shows a pose likelihood field in y-θ space with 50,000 sampled poses, given a measurement $^{\{V\}}Z$ and a template T; brighter color represent higher probability.*

If the algorithm has access to other sources of localization or motion information (e.g., control input, wheel odometry, steering angle sensor, GNSS, and visual odometry), the possible poses can be sampled from a distribution that is informed by these sources. In this work, visual odometry was used as an additional motion information source because of its accuracy and the fact that it requires no additional sensor hardware than the already available stereo camera. The visual odometry provided an incremental pose change from the current time step to the next time step (Cvišic et al., 2017), making it possible to sample in a more informed manner using the camera's (robot's) motion model. The Monte Carlo Localization algorithm with visual odometry is presented in **Algorithm 4**. The **SampleMotionModel()** function returns the multiple possible poses, given the odometry information.

**Algorithm 4 MonteCarloLocalization**(Visual odometry informed sampling)

**Input**: $^{\{C\}}Z_t$, a set of particles $\textbf{P}_{t-1}$, odometry measurement $u_t$, odometry noise $\Sigma_t$, **T**, rowRange
**Output**: $\hat{X}_t$, $\textbf{P}_t$, Pose covariance $\Omega_t$
**for** k = 1 : n **do**
   $< X_{t-1}^k, w_{t-1}^k > = P_{t-1}^k$
   $X_t^k = \textbf{SampleMotionModel}(u_t, \Sigma_t, X_{t-1}^k)$

```
    w_t^k = MeasurementModel( {C}Z_t, T, X_t^k, rowRange) # measurement step
    P_t^k = < X_t^k, w_t^k >
end for
i = argmax w_t^k
    0≤k≤1
X̂_t = X_t^i
Ω_t = CalculateCovarianceMatrix(X_t)
P_t = Resample(P_t)  # resampling step
return X̂_t, P_t, Ω_t
```

Our implementation of the Monte-Carlo algorithm (with uniform or visual odometry informed sampling) also estimates the covariance matrix of the estimated lateral offset and yaw by sampling 1% of the particles with the highest weights before resampling and calculating their covariance around the best pose estimate. The intuition behind this approach is that if 1% of best candidate particles are concentrated around the best pose estimate, this best pose estimate is more likely to be accurate, and the solution uncertainty (variance) is small. If 1% of the best candidate particles are spread out, the quality of the best pose guess tends to be low.

## 4   Experimental Design

The goals of the experiments were to evaluate the accuracy of the template-based localization method in different orchards and seasons (section 5.1); to evaluate the robustness of its accuracy against different template instances (section 5.2), and against mismatches between a template and traversed rows due to gaps in the tree rows (from missing or smaller trees) (section 5.3), and examine the localization accuracy as the sensor reaches the end of the row (section 5.4). Finally, the effect of the number of measurements used to build the template on the accuracy was investigated (section 5.7).

Experiments were conducted using a 3D camera in several rows, in a vineyard (L. *Vitis vinifera*) in spring, and in an apricot orchard (L. *Prunus armeniaca*), in different seasons. The metrics used to evaluate the accuracy of the lateral offset and heading with respect to the row centerline were the mean absolute error (MAE), the standard deviation (SD) of the absolute error, and the 95[th] percentile of the absolute error. Next, the experimental platform, the experimental design, and the ground truth generation process are presented in detail.

## 4.1 Experimental platform

The sensor used was a low-cost ZED stereo camera (Stereolabs Inc, San Francisco, CA). The field of view of this sensor is 90° (H) x 60° (V) x 100° (D) max, and the baseline of the stereo camera pair is 120 mm. It can output point clouds produced by stereo triangulation at 30 frames per second with 1080P resolution, using an NVIDIA GPU. Based on the specification datasheet, the camera's depth accuracy decreases quadratically over distance, and is better than 2% up to 3 m, and better than 4% up to 15 m. Of course, factors such as ambient light and object texture may affect the nominal accuracy. The camera is pre-calibrated and according to the company, does not require re-calibration. It also provides visual odometry at the same rate and supports communication via ROS (Robot Operating System). Our localization method can work with one single 3D camera without other sensors, which largely simplifies the overall system complexity and reduces the cost. If additional sensors are available, such as IMU (Inertial Measurement Unit) and wheel odometry, they can also be integrated with the template-based measurement model, and provide more informed sampling in the Monte Carlo localization framework.

A locally developed mobile robot was used as a mobile platform for data collection (Figure 11). The ZED stereo camera was mounted in the front center of the robot, facing forward. Two RTK-GNSS receivers provided ground truth for the position and heading in the vineyard. Ground truth in the orchard was measured using a different approach (see Section 3) because GNSS signals or RTK corrections were not available due to the foliage of large trees.

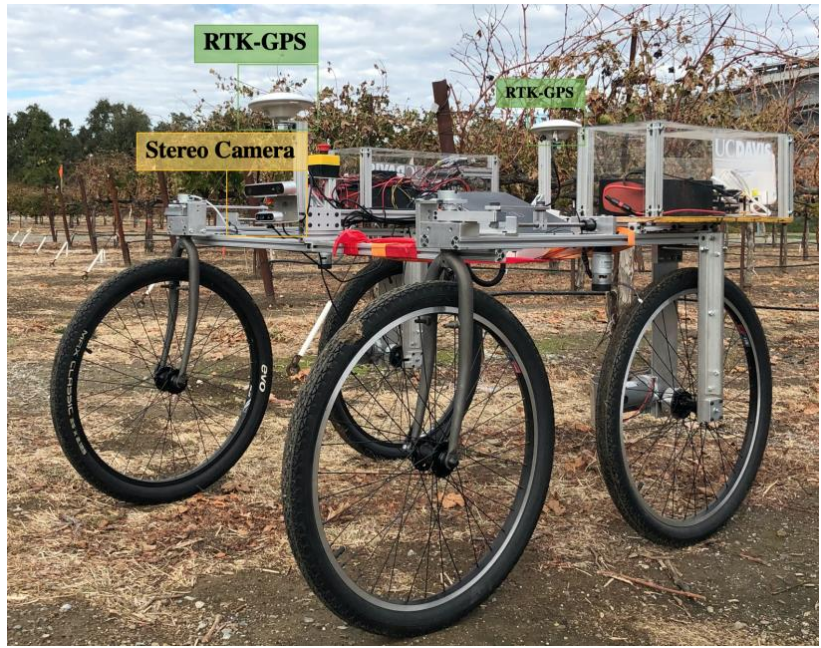

*Figure 11 The mobile robot that was used as our experimental platform.*

In all experiments, the robot traveled at a speed of 1 m/s. While traveling inside each row (vineyard or orchard), the robot was controlled remotely and steered to travel on a sinusoidally-shaped path, in order to sample the widest possible (collision-free) ranges for offset and yaw deviations from the centerline.

## 4.2 Experiments in a vineyard

Localization experiments were performed in an experimental vineyard at Davis, California, during the spring season, 2019 (Figure 12a). Ten random rows were traversed with the robot platform. The vineyard rows were straight; average row spacing was 3 meters; the height of the plants was 2.2 meters on average, and vines were planted every 1.8 meters. The trellis system of this vineyard is "vertical shoot position" which is a common and widely used trellis system. The vineyard was overall well-managed, with uniform tree type and spacing, and mowed rows on flat ground. Canopy management was relatively uniform, although there were branches/twigs extending inside the row, which registered in the point cloud but were not rigid or long enough to block the robot's travel. Also, some vines were missing, and left gaps in the row. The UTM (Universal Traverse Mercator) coordinates of the endpoints of ten vine rows were measured with an RTK GPS receiver; each row was approximately 90 meters long. Ground truth for the position and orientation of each row's centerline was computed from the measured row endpoints.

## 4.3 Experiments in a tree orchard

Localization experiments were also performed inside two rows of apricots trees, at Davis, California during 2018 and 2019. The apricots rows were straight, and the average row spacing was 5 meters; and trees were planted on flat ground, every 2.5 meters, on average. Each row was approximately 50 meters long. The apricot orchard was overall well-managed, with uniform tree type, age/size and spacing. Very light mannual pruning was performed. Experiments included traversal of the rows in the winter when trees were dormant and had no foliage (Figure 12b); in the summer, with dense foliage (Figure 12c), and in the spring with sparse foliage (Figure 12d).

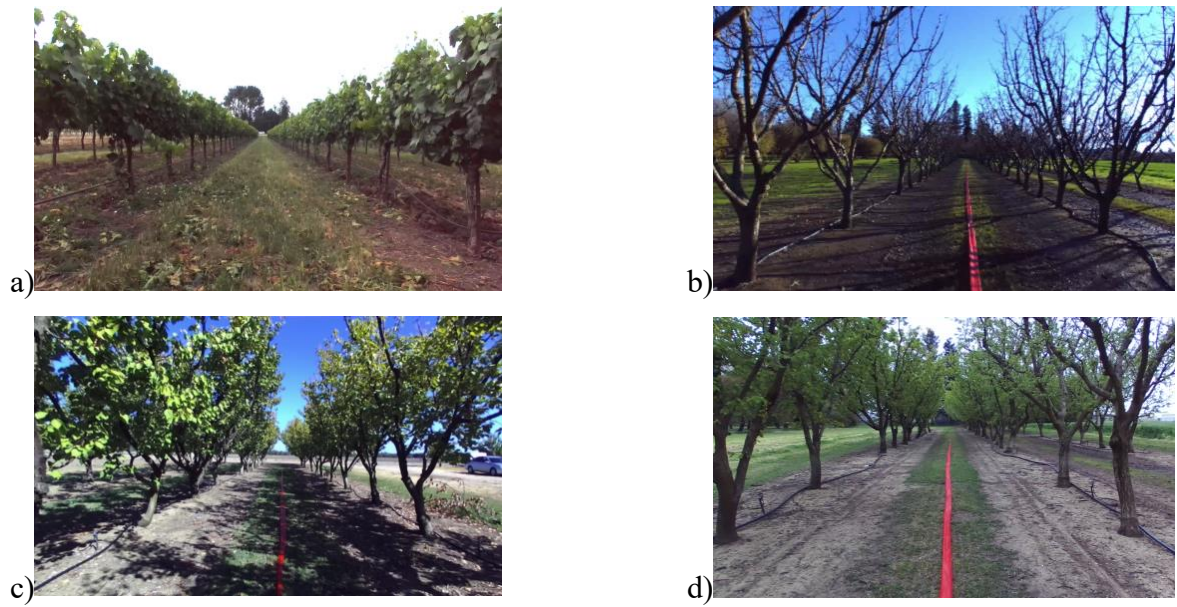

*Figure 12 a) Appearance of the vineyard in spring; appearance of the apricot tree orchard in b) winter, c) summer, and d) spring.*

Inside the tree rows, the RTK-GNSS did not provide reliable localization because of tall canopies and foliage. As an alternative way to generate the ground truth for the sensor's offset and yaw with respect to the row's centerline, a physical centerline was used. A visually prominent colored rope was placed – and stretched - along the center of the orchard row (Figure 13a), and an image processing pipeline was developed to detect and localize it. The pipeline included three steps; 1) detect the rope as a line in the camera's image space (Figure 13b); 2) project the detected line back onto the physical ground plane in the camera frame, using a well-calibrated camera projection matrix (Figure 13c); 3) compute the lateral offset and heading angle of the camera relative to the rope-defined centerline. The template-based localization algorithm does not use any color information, so this colored rope did not affect the algorithm's performance. We pre-evaluated the localization accuracy of this rope-based method in a field where RTK-GNNS was available. The y-offset difference was $0 \pm 0.012(SD)$ m, and the heading difference was $0 \pm 0.00043(SD)$ rads between the rope-based method and RTK-GNSS.

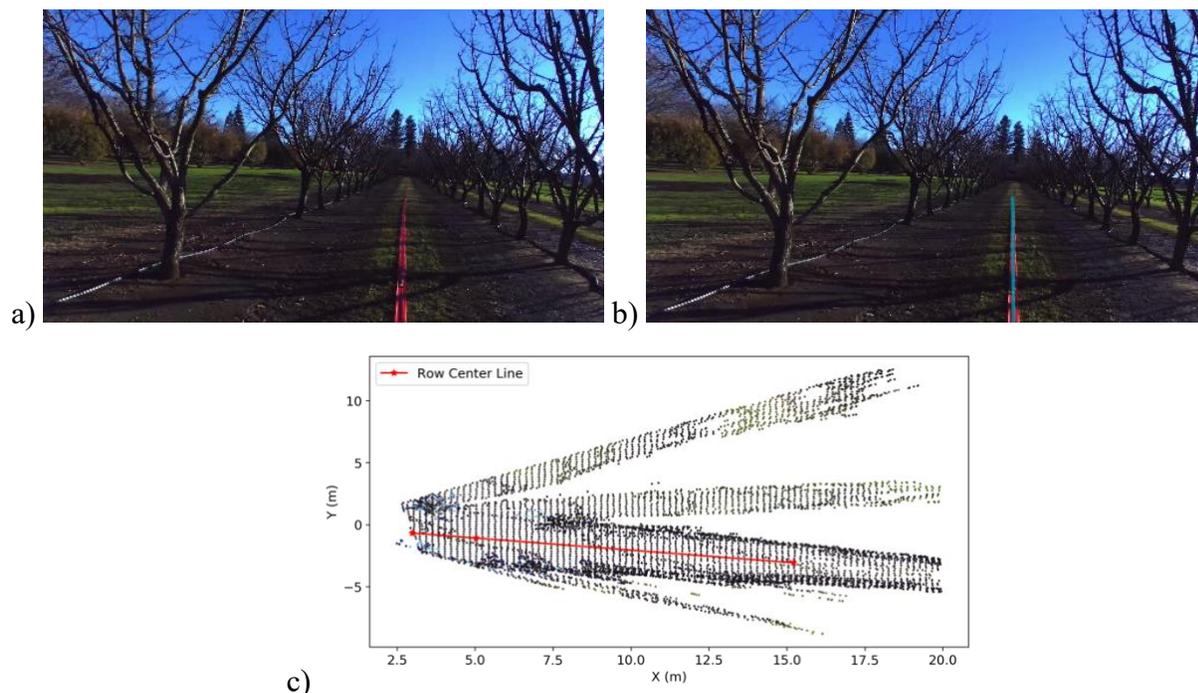

Figure 13 a) The figure shows the red rope in camera's image view; b) the blue line is the extracted line in the image space; c) the red line is the extracted line projected onto the ground plane (top-down view of the corresponding point cloud).

## 4.4 Autonomous control experiment

To demonstrate that the template-based localization method can provide accurate enough localization results for autonomous guidance in orchard rows, an autonomous navigation experiment was performed in a vineyard at Davis, California, in Summer 2020. A simple proportional line following controller was used for both heading and offset (Corke, 2017). The turning at the end of the row was initiated and guided by the RTK-GNSS, whereas the in-row navigation relied solely on the template-based localization. Localization was based on the measurement model without the particle filter to remove the effect of visual odometry. In practice, adding a particle filter would provide better localization accuracy. RTK-GNSS trajectories were recorded during the experiment to assess the tracking errors.

## 4.5 Comparison with existing methods

The template-based localization method was compared against two existing in-row navigation methods, which we refer to as 'baseline' methods. Since our method is pointcloud-based and does not rely on specific features (e.g., tree trunk, skyline, irrigation lines), the baseline methods we selected are also based on point clouds as their input and do not rely on specific features.

### 4.5.1 Selected baseline methods

There is no open-source code available for the two baseline methods we selected, so we reimplemented them, made necessary improvements, and tried our best to tune the related parameters. The pseudocodes of our implementation - and modification details - of the baseline methods are provided in the appendix. In the original papers, the localization results of the methods were filtered with odometry information using the Extended Kalman Filter (EKF), similarly to our method that uses a particle filter. However, the purpose of the comparison is to compare the localization methods as 'stand-alone'. Therefore, all the results generated and reported here (including those of our method) did not involve any filtering.

#### 4.5.1.1 Baseline1 algorithm

Stefas et al., (2016) proposed a binocular depth-based navigation algorithm that localizes the robot by 1) using RANSAC to estimate the x-y ground plane; 2) projecting the non-ground points into the x-y plane, and 3) fitting two tree row lines using RANSAC line fitting. We call this method the "baseline1" method.

#### 4.5.1.2 Baseline2 algorithm

Zhang et al., (2013) proposed an algorithm for localization inside orchard rows that uses 3D point clouds. The method spits the point cloud into a left and a right point set, and repeatedly selects two sets of random points from the left and right point sets to calculate pairs of candidate parallel lines. The parallel lines with the minimum mean squared distance to the inlier points are selected as the tree row lines. The lateral error/offset from the centerline was further refined using point density information from the point cloud. We call this method the "baseline2" method.

### 4.5.1.3 Comparison experiments

We compared our method with the baseline1 and baseline2 methods, and the baseline2 method with offset refinement on all the data we collected. For all the experiments, we applied a spatial cutoff filter (shaped as a rectangular box) to the point cloud to filter. This filter removed all points higher than 2.5m from the ground and 0 m below the ground, because the measurement points from the ZED camera contain a lot of noise above 2.5 m, which affects significantly the baseline methods. (The noise does not affect the template-based method too much, but for a fair comparison, we applied the same cutoff filter to all methods, in all experiments). The y-range of the cutoff filter was -5m to 5m, and the x-range of the cutoff filter was 0 m to 20 m.

The template-based method and the baseline methods were compared as follows:
1. We compared the mean absolute error (MAE), the standard deviation (SD) of the absolute error, and the 95th percentile of the absolute error of the estimated lateral and heading deviations relative to the row centerline.
2. We compared the error distributions of the accumulated error of the estimates of the lateral offset and heading.

The above comparisons were performed when the robot's heading relative to the centerline was reasonably small (less than 0.3 rads/17.2º). The reason is that during autonomous row traversal, under normal operating conditions, the robot is not expected to deviate too much from the centerline. However, the localization robustness when the heading deviation is large (e.g., robot faces more toward the side of the row) is very important for unexpected situations when autonomous guidance fails, and the system must perform failure recovery. Therefore, we also compared the algorithms' accuracies at "large heading deviations" (> 0.3 rads) separately.

# 5 Experimental Results

## 5.1 Localization accuracy

We refer to a specific orchard at a specific season as an "operational scenario". An orchard row template is valid for an entire operational scenario. All the experimental results were generated using templates built by 100 consecutive point-cloud measurements (which corresponds to approximately 7 seconds or 7 meters of data) for each operational scenario, as the sensor moved inside a row. Figure 14 shows the middle slices of the 3D templates generated for our experiments.

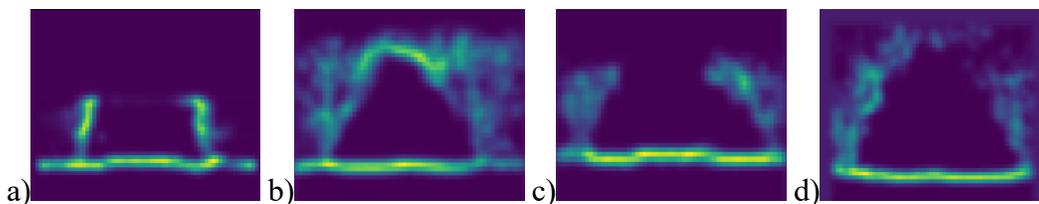

*Figure 14 Visualization of the middle slice of templates generated in a vineyard in spring (a) and in an apricot tree orchard in winter (b), summer (c), and spring (d).*

For all experiments, we ran our algorithm offline with and without visual odometry information. Without visual odometry, our particle filter sampled poses from a uniform distribution in the $\{T\}$ frame ( $y \in U[-0.8, 0.8]$ m, $\theta \in U[-0.6, 0.6]$ rads). With visual odometry, the Monte-Carlo sampling procedure described in **Algorithm 4** was used. An example of the final localization output in a vineyard row with visual odometry is shown in Figure 15.

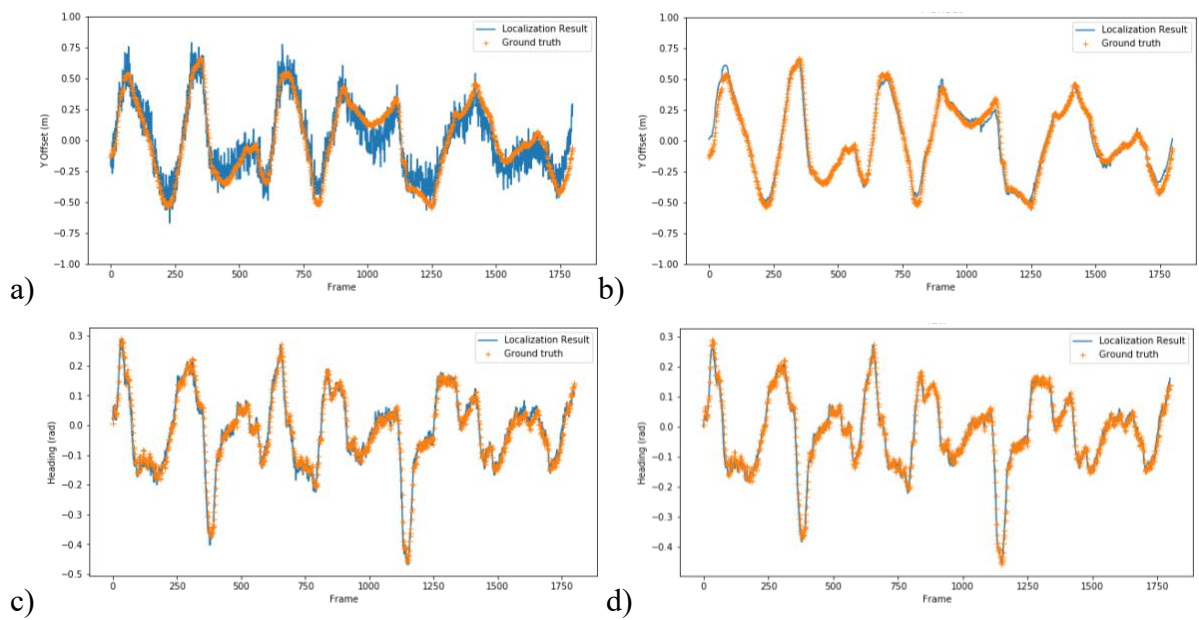

*Figure 15 The localization results of lateral offset (a) and the heading (c) with uniform sampling are overlaid with ground truth. The localization results of the lateral offset (b) and heading (d) with visual odometry informed sampling are overlaid with ground truth.*

The overall localization results for all operational scenarios and also for each run are reported in Table 1.

| Operational Scenario | Visual Odometry Informed Sampling | | | | | | | | Uniform Sampling | | | | | | | |
|---|---|---|---|---|---|---|---|---|---|---|---|---|---|---|---|---|
| | Y error (meter) | | | | | Yaw error (rad) | | | Y error (meter) | | | | | Yaw error (rad) | | |
| | MAE† | SD† | 95%† | MAE / Row spacing† | 95% / Row spacing | MAE | SD | 95% | MAE | SD | 95% | MAE / Row spacing | 95% / Row spacing | MAE | SD | 95% |
| Vineyard | 0.03 | 0.02 | 0.07 | 1.0% | 2.5% | 0.01 | 0.01 | 0.04 | 0.10 | 0.07 | 0.24 | 3.2% | 8.0% | 0.02 | 0.01 | 0.04 |
| Apricots Winter | 0.09 | 0.08 | 0.21 | 1.8% | 4.3% | 0.03 | 0.03 | 0.07 | 0.17 | 0.17 | 0.47 | 3.3% | 9.4% | 0.02 | 0.03 | 0.04 |
| Apricots Spring | 0.07 | 0.06 | 0.20 | 1.5% | 4.1% | 0.02 | 0.01 | 0.04 | 0.18 | 0.14 | 0.46 | 3.5% | 9.2% | 0.02 | 0.04 | 0.04 |
| Apricots Summer | 0.09 | 0.06 | 0.20 | 1.8% | 4.0% | 0.02 | 0.02 | 0.05 | 0.18 | 0.14 | 0.46 | 3.6% | 9.2% | 0.02 | 0.02 | 0.05 |

† MAE: Mean Absolute Error;   SD: Standard Deviation;   95%: 95th Percentile;
† † The average row spacing is 3 meters for the vineyard and 5 meters the apricots orchard

*Table 1 Localization results with and without visual odometry, for each scenario. The average row spacing is 3 meters the vineyard and 5 meters for the apricots orchard.*

In all the operational scenarios, our method localized the vehicle with heading MAE below 0.03 rad and lateral MAE below 5% of the row spacing, without visual odometry. When visual odometry was used for informed sampling, the lateral MAE dropped below 2% of the row spacing. The results were consistent across different orchards and seasons.

In all the experiments we applied a cutoff filter with a range of 0 to 20 m in the x-direction, -5 to 5 m in the y-direction, and 0 to 4 m in the z-direction to get rid of measurement points not in the current row to improve the localization accuracy. However, the effect of the cutoff filter is marginal, we also evaluated the uniform sampling localization results in the vineyard without applying any cutoff filter and the results are shown in Table 2. As we can see from the results, using a cutoff filter can improve the localization accuracy but the difference between using and not using a cutoff filter is very small.

| Vineyard | Y error (meter) | | | Yaw error (rad) | | |
|---|---|---|---|---|---|---|
| | MAE | SD | 95% | MAE | SD | 95% |
| With cutoff filter | 0.10 | 0.07 | 0.24 | 0.02 | 0.01 | 0.04 |
| Without cutoff filter | 0.11 | 0.09 | 0.28 | 0.02 | 0.01 | 0.04 |

*Table 2 Uniform sampling localization results in the vineyard, with and without a cutoff filter*

## 5.2 Localization robustness against template instance

An important assumption in the proposed method is that a template built from a set of data from one or more rows can be used for localization in all rows – in the same orchard - without significant loss in localization accuracy. To evaluate the validity of this assumption, a template instance that was generated using 100 consecutive measurements while moving inside row $k$ (for each $k = 0, 1, \ldots, 9$) was used to localize the robot – with and without visual odometry informed sampling - in evaluation runs inside all the ten rows in the vineyard block. The localization MAEs - with and without visual odometry - are given in Figure 16a and Figure 16b, respectively.

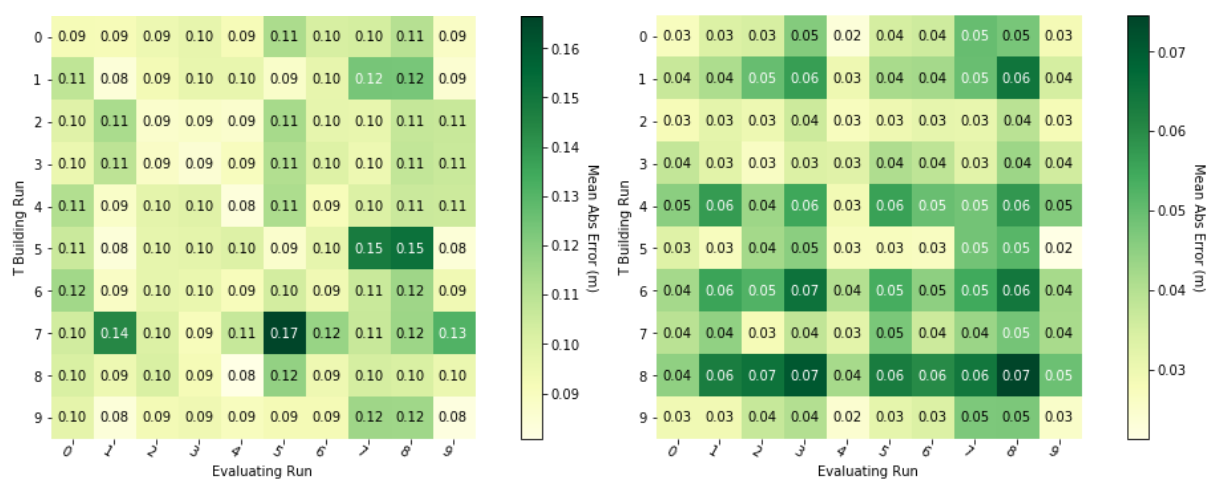

*Figure 16 Localization errors (MAE) when a template generated using data from row k was used to localize the robot in an evaluation run in row j (0≤ k, j ≤ 9); lighter colors correspond to better accuracy. a) Left: Monte-Carlo localization with uniform sampling. b) Right: Monte-Carlo localization with informed sampling from visual odometry.*

The elements $(k, k)$ on the main diagonals corresponded to localization errors when the template that was built with data from row $k$ was used for localization inside the same row $k$. When using uniform sampling, the maximum MAE was 0.11 m, whereas the off-diagonal MAE was 0.17 m. However, when using visual odometry informed sampling, the maximum MAE was 0.07 m for both the on and off-diagonal elements. Also, some templates resulted in overall slightly better results than others (e.g., in the right matrix, row #2 is much lighter-colored than row #8), Table 3 presents detailed localization errors when all vineyard rows were traversed using a template based on measurements from row #2.

| Run ID | Visual Odometry Informed Sampling | Uniform Sampling |
|---|---|---|

|   | Y error (meter) | | | | | Yaw error (rad) | | | Y error (meter) | | | | | Yaw error (rad) | | |
|---|---|---|---|---|---|---|---|---|---|---|---|---|---|---|---|---|
|   | MAE† | SD† | 95%† | MAE / Row spacing† | 95% / Row spacing | MAE | SD | 95% | MAE | SD | 95% | MAE / Row spacing | 95% / Row spacing | MAE | SD | 95% |
| 0 | 0.03 | 0.02 | 0.07 | 0.9% | 2.5% | 0.01 | 0.01 | 0.04 | 0.10 | 0.07 | 0.24 | 3.2% | 7.9% | 0.02 | 0.01 | 0.04 |
| 1 | 0.03 | 0.02 | 0.08 | 1.1% | 2.6% | 0.02 | 0.01 | 0.04 | 0.11 | 0.08 | 0.26 | 3.5% | 8.8% | 0.02 | 0.01 | 0.05 |
| 2† | 0.03 | 0.03 | 0.08 | 1.0% | 2.7% | 0.02 | 0.01 | 0.04 | 0.08 | 0.07 | 0.21 | 2.7% | 6.9% | 0.02 | 0.01 | 0.04 |
| 3 | 0.05 | 0.03 | 0.11 | 1.6% | 3.8% | 0.02 | 0.02 | 0.05 | 0.08 | 0.06 | 0.20 | 2.7% | 6.7% | 0.02 | 0.02 | 0.05 |
| 4 | 0.03 | 0.03 | 0.08 | 1.0% | 2.7% | 0.02 | 0.02 | 0.05 | 0.08 | 0.07 | 0.21 | 2.8% | 7.1% | 0.02 | 0.02 | 0.05 |
| 5 | 0.03 | 0.02 | 0.07 | 1.1% | 2.5% | 0.01 | 0.01 | 0.03 | 0.11 | 0.09 | 0.28 | 3.6% | 9.2% | 0.01 | 0.01 | 0.03 |
| 6 | 0.03 | 0.03 | 0.08 | 1.2% | 2.8% | 0.01 | 0.01 | 0.03 | 0.10 | 0.08 | 0.25 | 3.2% | 8.3% | 0.02 | 0.01 | 0.04 |
| 7 | 0.04 | 0.03 | 0.10 | 1.4% | 3.3% | 0.02 | 0.02 | 0.05 | 0.10 | 0.08 | 0.24 | 3.3% | 8.1% | 0.02 | 0.01 | 0.05 |
| 8 | 0.04 | 0.03 | 0.11 | 1.4% | 3.6% | 0.02 | 0.02 | 0.05 | 0.10 | 0.08 | 0.25 | 3.5% | 8.4% | 0.02 | 0.01 | 0.05 |
| 9 | 0.03 | 0.02 | 0.06 | 0.8% | 2.0% | 0.01 | 0.01 | 0.03 | 0.10 | 0.08 | 0.25 | 3.3% | 8.4% | 0.02 | 0.01 | 0.04 |

† MAE: Mean Absolute Error;   SD: Standard Deviation;   95%: 95 Percentile;
† The average row spacing is 3 meters for the vineyard.
† Template is built using data from run id 2

*Table 3 The table shows accuracy results when a template that was built from measurements in row #2 of the vineyard is used for localization in all ten rows of the vineyard.*

Overall, the above results suggested that a template developed from one row could be used for localization in other rows without significant loss of accuracy.

## 5.3 Localization robustness against gaps in rows

The template-based localization method is based on the assumption that when the 3D sensor is on the center line and aligned to it, the spatial distribution of the point cloud sensed anywhere along an orchard row matches the point cloud distribution of the template. However, in commercial orchards, it is very common that one or more trees are missing (e.g., due to disease) or are much smaller (because of replanting), thus creating "gaps" along the tree lines. Such gaps represent extreme cases/outliers of variability inside a row. Examples of missing trees from our data can be seen in Figure 17a. In the left image, one tree is missing on the left of the row, and in the right image, one tree is missing on both sides of the row.  Figure 17b shows the top-down views of the corresponding 3D point clouds (sliced at 1.5 m height), transformed into the template frame $\{T\}$.

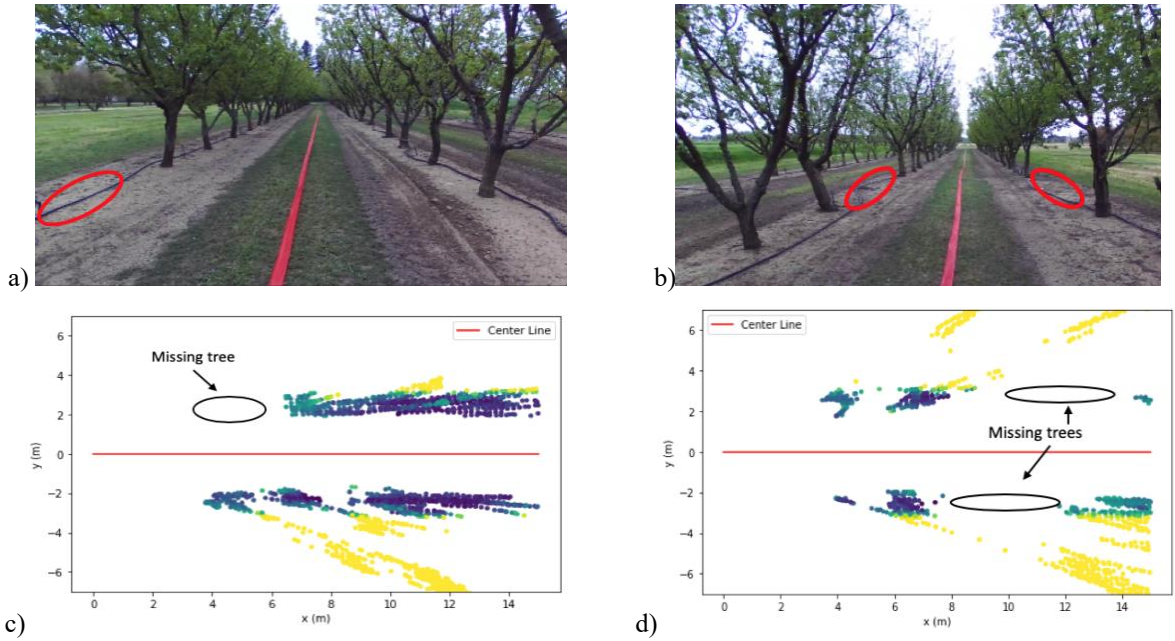

*Figure 17 a, b) Camera views of orchard rows with gaps (missing trees) (red ellipses). c, b) Top-down view of the point cloud (excluding treetop and ground); black ellipses are gaps.*

To evaluate the robustness of the template method in the presence of gaps, sets of points in the measurements were artificially removed, to simulate such gaps. A length of 1 m was used as a "unit length" for gaps in the point cloud data; this length is referred to as a "unit-tree." Smaller trees could result in one unit missing, whereas larger missing trees could result in more than one consecutive missing units. Since the 3D sensor used in this work had a range of 20 m, each side of the measurement was split into 20 units, as shown in Figure 18a, where each color corresponds to one unit-tree. Then, $n$ unit-trees were randomly removed from the measurement, and the localization error was evaluated using the remaining measurement points (e.g., green points in Figure 18b).

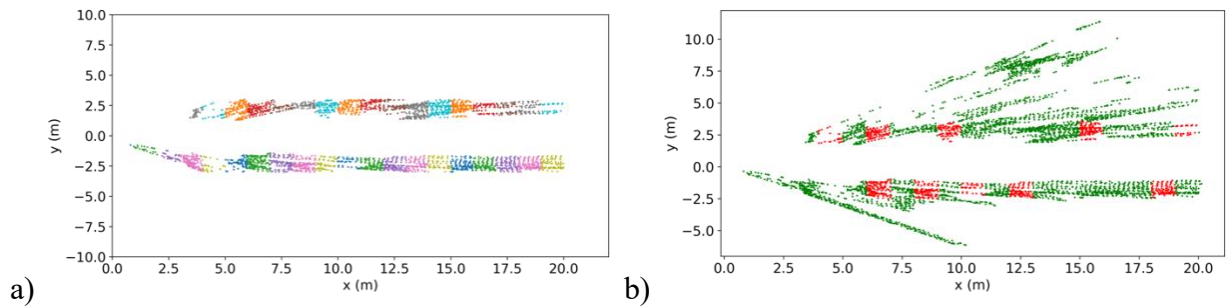

*Figure 18 a) This figure shows unit-trees in a measurement; each color represents a unit-tree. b) A point-cloud example when four random unit-trees are removed from each side of the row; green points represent the remaining points.*

There are C(40, n) different combinations for removing *n* unit trees from 40 unit trees. Given the very large number of possible combinations, 100 were sampled randomly to evaluate the performance of the approach, at a given number of gap units *n*, as *n* was increased from 0 to 40. The lateral offset and heading error results for the vineyard data are shown in Figure 19, and for the apricot orchard data, in Figure 20.

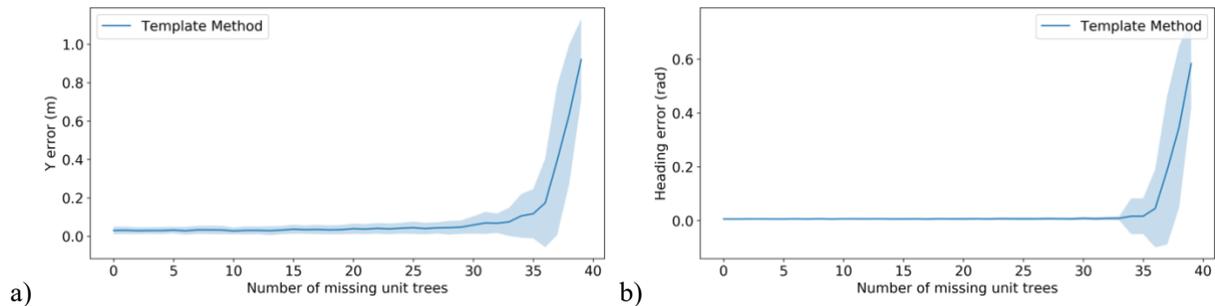

*Figure 19 a) Offset error curve vs. the number of missing unit-trees in the vineyard. b) Heading error curve as a function of the number of missing unit-trees in the vineyard. The shaded region indicates the area of error standard deviation.*

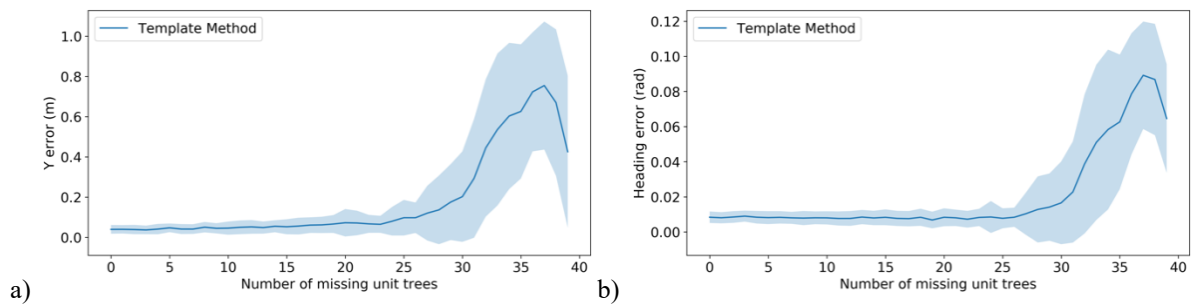

*Figure 20 a) Offset error curve vs. the number of missing unit-trees in apricot orchard. b) Heading error curve as a function of the number of missing unit trees in the apricot orchard. The shaded region indicates the area of error standard deviation.*

The localization accuracy remained almost constant until a certain threshold-number of unit trees were removed from the measurements. After inserting more gaps than this threshold, the errors grew very fast. In the particular vineyard, the threshold was approximately 30 unit trees (out of 40), i.e., the proposed localization approach performed robustly as long as no more than 75% of the measurement points were missing. For the apricot orchard, this threshold was approximately 25 (out of 40), i.e., the algorithm was robust as long as no more than 62.5% of the measurements were missing. The difference in algorithm robustness between the vineyard and the apricot orchard could be attributed to the fact that apricot trees were spaced farther apart from each other than grapevines. In real orchards and vineyards, the percentage of missing trees is very small.

Hence, the proposed localization method is not expected to encounter such situations, and its robustness against gaps in tree rows seems adequate.

Along with the errors, we also extracted the standard deviation of the offset and heading from the covariance matrix returned by our localization algorithm in the vineyard case (Figures 18 a, b, respectively). The standard deviations represent the *confidence* of the localizer. Our localization algorithm started reporting high standard deviation in the lateral offset and the heading at the same time when the localization errors grow. The magnitude of the standard deviation is highly correlated with the actual error, as shown in Figure 18 c, d). These results showed that our algorithm could correctly report the uncertainties in lateral offset and yaw. The reported uncertainties could be used by other robot software modules, such as Bayesian filters or fail-safe modules, which can act accordingly when the uncertainly becomes large.

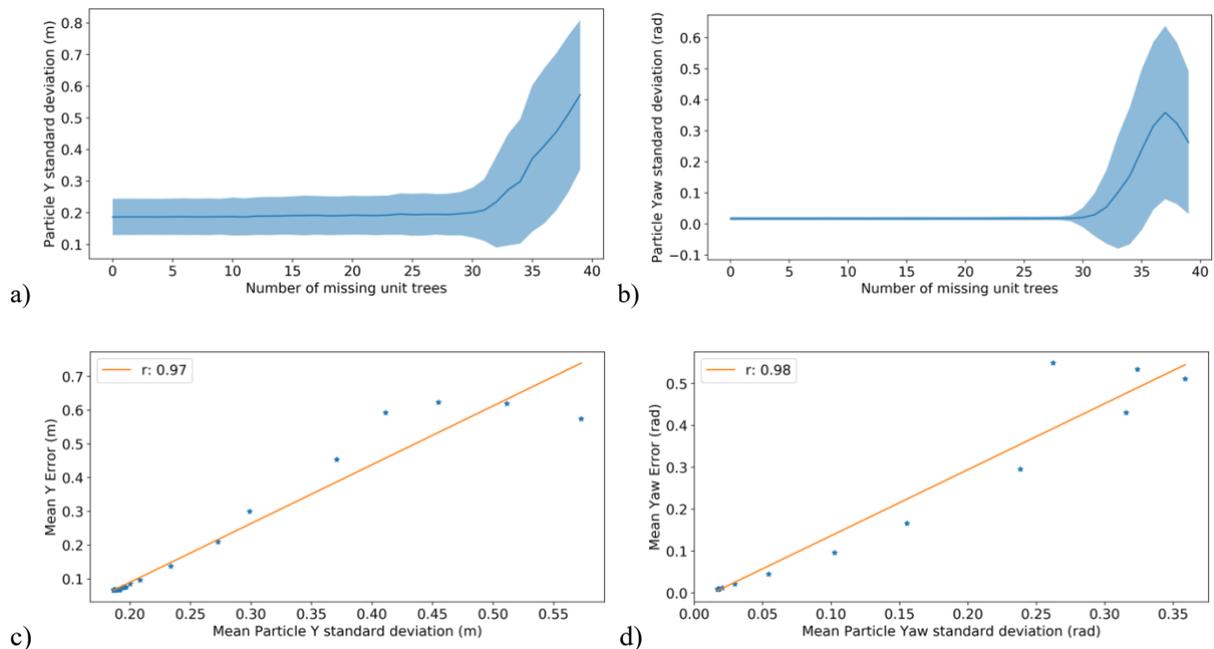

*Figure 21 a) Lateral offset standard deviation curve vs. the number of missing unit trees in the vineyard. b) Heading standard deviation curve vs. the number of missing unit trees in the vineyard. The shaded region indicates the area of error standard deviation. c) and d) show the relationship between standard deviations of sampled particle poses and localization errors.*

## 5.4 Localization accuracy near the end of the row

The template-based method is designed for in-row navigation, but it is important to understand the method's behavior when the vehicle approaches a row's end, in order to integrate this method in the future into a full orchard navigation system. A simulation experiment was designed to analyze the localization accuracy when the vehicle approaches the end of the row that it is currently traversing. The measurement points that were further than *d* meters away in the template frame's x-axis were removed. The remaining points represent the row end in *d* meters from the vehicle. An example of the simulated measurement is shown in Figure 22, and the localization accuracy as the vehicle "approaches the row end" is shown in Figure 23.

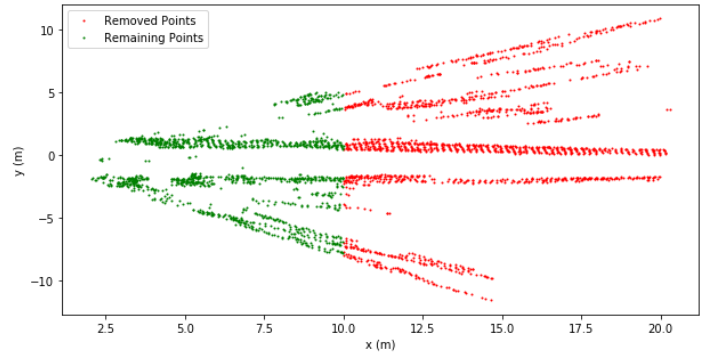

*Figure 22 This figure shows points farther from the sensor than d meters are removed to simulate row exiting. The red points are removed points, and the green points are remaining points.*

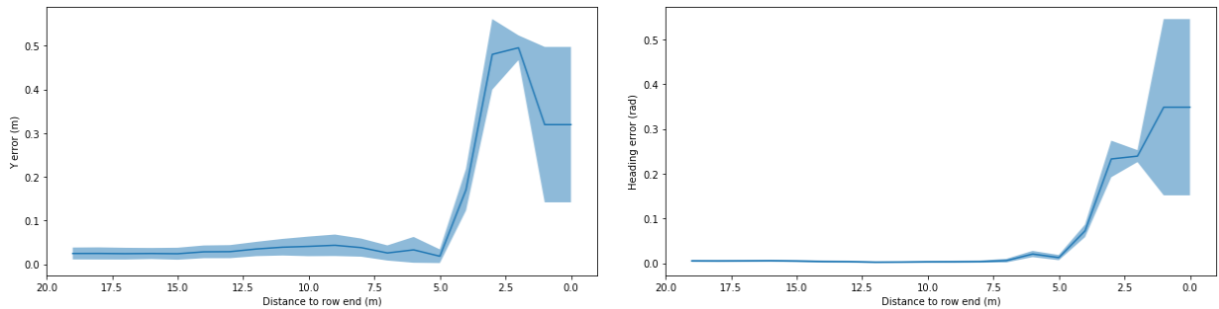

*Figure 23 These figures show the localization accuracy as a function of the robot distance to the end of the row. The left figure shows the lateral offset error (Y), and the right figure shows the heading error. The shaded region indicates the area of error standard deviation*

The row-exiting results indicate that the localization accuracy did not change much until the vehicle was 5 m away from the row end. The remaining observed measurement points when the vehicle was 5 m away are shown in Figure 24 a). The overall row-exiting results indicated that the method could work well for an in-row vehicle until 5 m away from the row-end. The specific number may change under different situations; however, we can see that the safety margin is significant. Also, when the vehicle gets too close to the row end, the localization algorithm reports high localization uncertainty.

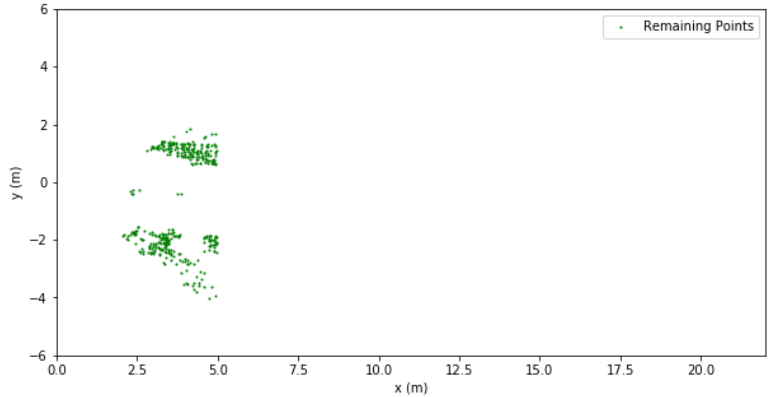

*Figure 24 The figure shows remaining measurement points after 75% of points were removed to simulate the robot is 5 m away from the row end.*

## 5.5  Localization performance inside curved rows

Most modern commercial orchards are established on flat ground, and RTK-GNSS or very accurate surveying laser equipment is used to plant the trees, resulting in perfectly straight rows. However, it is possible that some existing orchards have rows that are slightly curved due to the landscape or their imperfect establishment. In order to test the method's performance inside rows of increasing curvature, a set of simulation experiments were conducted. (Simulations were necessary, as it was not possible to find commercial orchards with varying curvatures, to conduct physical experiments.)

First, curved vineyard rows of increasing curvature, $\kappa$, were created, by bending the centerline of the existing straight rows in the vineyard, so that they "sat" on the circumference of circles of decreasing radius, $R$ ($\kappa=1/R$). The row's sensor point cloud measurements were bent accordingly. An example of a curved row on the circumference of a circle of 135m radius is shown in Figure 22.

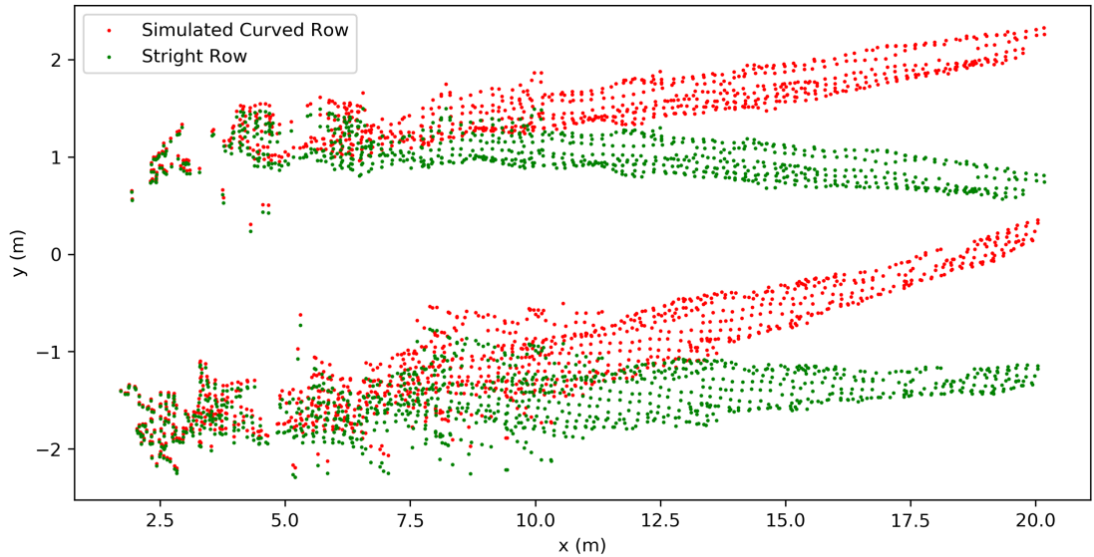

*Figure 25 An example of a simulated curved row and its corresponding measurement points (Top down view). The row's center axis is on a circle of radius R= 150m radius.*

For the evaluation, a template was built using a set of straight rows, and was used for localization inside the curved rows. Two different sensor ranges were used: 20 m (scenario 1) and 10 m (scenario 2). It was hypothesized that a longer range would result in larger errors, as the effect of curvature would be more pronounced over a longer distance. Next, the localization accuracy was assessed inside curved rows with radius ranging from 135 m to 500 m. The results are shown in Figure 26.

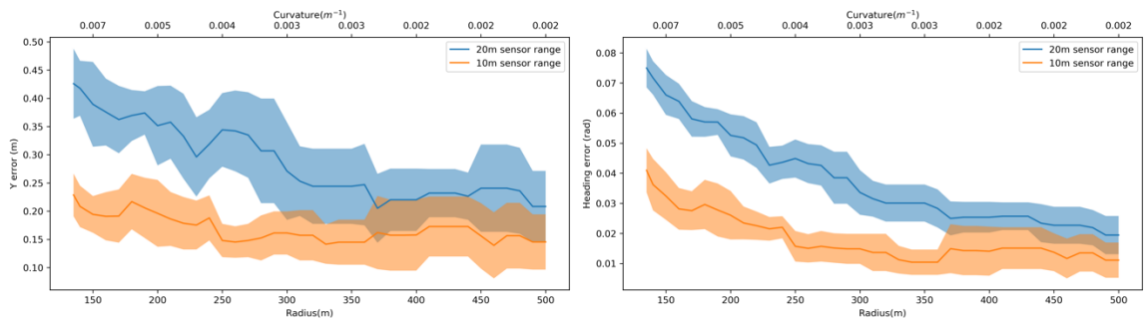

*Figure 26 These figures show the localization accuracy as a function of the orchard row's radius, when using a template built from straight rows, with sensor range at 10 m (scenario 1) and 20 m (scenario 2). The left figure shows the lateral offset error (Y), and the right figure shows the heading error. The shaded regions indicate the standard deviation of the error.*

As anticipated, the results show that when the template was built using data from straight rows, the lateral and orientation errors of the template-based localization inside curved rows increased as the curvature of the row increased. Using a shorter sensing range increased the localization robustness, i.e., the lateral and orientation error with a 20m-range increased faster the corresponding errors with a 10m-range. It is important to note that in real-world orchards, the radius of curvature is expected to be greater than 498 m (see Appendix for radius range). A t-test for R=500 m showed that the errors when using the two ranges were not statistically significant. Hence, in this experiment, the performance of the template method was shown not to be affected by small curvature in the orchard rows.

## 5.6 Template voxel size study

To explore the effects of the template resolution on localization accuracy, the template was built using a voxel size that ranged from 0.02 m to 1 m (equally for the x, y, and z directions). Localization performance was evaluated inside one vineyard row, using 100 consecutive point-cloud vineyard measurements. Figure 27 a) and b) show the localization accuracy with respect to the template voxel size. One can see that the localization error was minimum when the voxel size was 0.1 m. As the voxel size increased beyond 0.1 m - and template resolution worsened - the localization error increased. However, below 0.1 m, when the voxel size decreased the localization error increased again. A possible reason is that when the voxel size is too small, the occupancy *frequency* of each voxel becomes very small and is affected more by noise in the measurements; hence, the estimates of the occupancy probability and the likelihood field become worse. Also, a finer template will result in a larger template storage size, since the storage size of the template is linear to the number of voxels, as shown in Figure 27 c). Based on the localization accuracy and storage size, using a template with 0.1 m size for the voxels was the best choice for the tested environments (which were not different from most commercial orchards and vineyards).

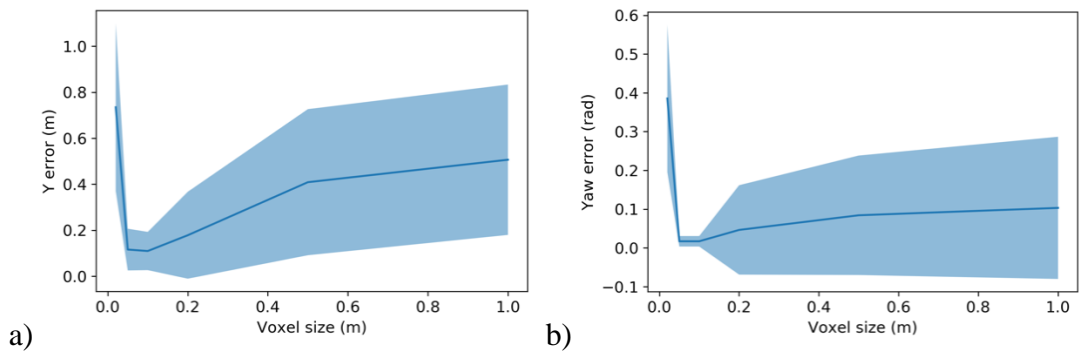

a) b)

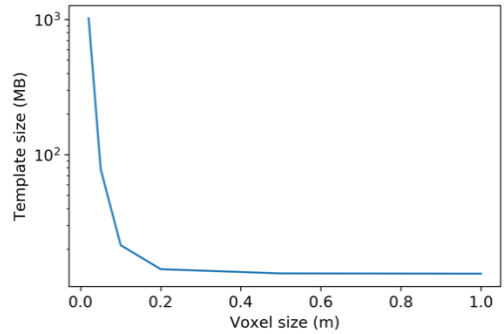

c)

*Figure 27 a) Localization y error with respect to the template voxel size. b) Localization yaw error with respect to the template voxel size. c) template storage size with respect to the template voxel size.*

## 5.7 Template data requirements

Given that a template should capture the "expected" or "typical" structure of the orchard row (by storing spatial occupancy probabilities), one important question is, how many measurements are needed to build a good template. To explore this question, seven templates were built using 1, 5, 10, 20, 100, 200, and 300 point-clouds from consecutive frames of data in the vineyard. Then, the localization results - when each one of the seven templates was used - were evaluated on all vineyard data (Table 4).

| Number of Measurements | With visual odometry | | | | | | | | Without visual odometry | | | | | | | |
|---|---|---|---|---|---|---|---|---|---|---|---|---|---|---|---|---|
| | Y error (meter) | | | | | Yaw error (rad) | | | Y error (meter) | | | | | Yaw error (rad) | | |
| | MAE† | SD† | 95%† | MAE / Row spacing† | 95% / Row spacing | MAE | SD | 95% | MAE | SD | 95% | MAE / Row spacing | 95% / Row spacing | MAE | SD | 95% |
| 1 | 0.03 | 0.03 | 0.08 | 1.00% | 2.80% | 0.01 | 0.01 | 0.04 | 0.11 | 0.09 | 0.29 | 3.80% | 9.50% | 0.02 | 0.01 | 0.04 |
| 5 | 0.03 | 0.03 | 0.08 | 1.00% | 2.80% | 0.01 | 0.01 | 0.04 | 0.11 | 0.09 | 0.28 | 3.70% | 9.30% | 0.02 | 0.01 | 0.04 |
| 10 | 0.03 | 0.02 | 0.07 | 1.00% | 2.40% | 0.01 | 0.01 | 0.04 | 0.11 | 0.09 | 0.28 | 3.70% | 9.30% | 0.02 | 0.01 | 0.04 |
| 20 | 0.03 | 0.02 | 0.07 | 1.00% | 2.20% | 0.01 | 0.01 | 0.04 | 0.11 | 0.08 | 0.26 | 3.60% | 8.70% | 0.02 | 0.01 | 0.04 |
| 100 | 0.03 | 0.02 | 0.07 | 1.00% | 2.50% | 0.01 | 0.01 | 0.04 | 0.1 | 0.07 | 0.24 | 3.20% | 8.00% | 0.02 | 0.01 | 0.04 |
| 200 | 0.03 | 0.02 | 0.07 | 0.90% | 2.30% | 0.01 | 0.01 | 0.04 | 0.1 | 0.08 | 0.25 | 3.30% | 8.20% | 0.02 | 0.01 | 0.04 |
| 300 | 0.03 | 0.02 | 0.07 | 1.10% | 2.50% | 0.01 | 0.01 | 0.04 | 0.1 | 0.08 | 0.25 | 3.40% | 8.30% | 0.02 | 0.01 | 0.04 |

† MAE: Mean Absolute Error;    SD: Standard Deviation;    95%: 95 Percentile;

† The average row spacing is 3 meters for this vineyard.

*Table 4 Localization results in the vineyard, using templates built with different number of measurements*

The results in Table 4 show that even a small number of measurements could result in a good template. The mean absolute error did not change much, and the 95% percentile of the Y error decreased slightly as the number of measurements increased. After a certain number of measurements, the improvement saturated; this number was 100 in our vineyard experiments.

## 5.8  Autonomous control experiment

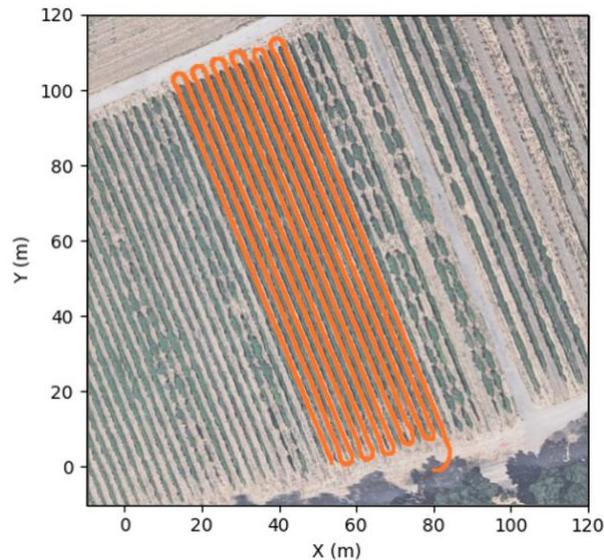

*Figure 28 Autonomous navigation in orchard trajectory*

The RTK-GNSS trajectory of the autonomously controlled experiment is shown in Figure 28. The robot traveled inside the rows and descriptive statistics of the in-row tracking error are shown in Table 5. Of course, the tracking error is the combined result of the line following controller and the template-based localization method. A more advanced line following controller can improve the tracking error.

|  | MAE | SD | 95% |
|---|---|---|---|
| Offset tracking error (m) | 0.10 | 0.065 | 0.22 |
| Heading tracking error (rad) | 0.017 | 0.017 | 0.04 |

*Table 5 Autonomous navigation in-row tracking error in autonomous navigation*

## 5.9 Comparison with other methods

The overall localization results are shown in Table 6. One can see that the template-based method's accuracy surpassed both baseline methods at all experiments. (Note that all the measurement points higher than 2.5m from the ground are cutoff in this experiment for a fair comparison, so the template method results are slightly different from Table 3.)

### 5.9.1 Overall accuracy comparison

| Method | Y error (meter) | | | Yaw error (rad) | | |
|---|---|---|---|---|---|---|
|  | MAE† | SD† | 95%† | MAE | SD | 95% |
| | | | Vineyard | | | |
| Baseline1 | 0.20 | 0.37 | 0.45 | 0.02 | 0.02 | 0.05 |
| Baseline2 | 0.36 | 0.39 | 1.26 | 0.03 | 0.05 | 0.07 |
| Baseline2 refined | 0.26 | 0.23 | 0.69 | 0.03 | 0.05 | 0.07 |
| Template method | **0.12** | **0.09** | **0.29** | **0.02** | **0.02** | **0.04** |
| | | | Apricot Spring | | | |
| Baseline1 | 0.45 | 0.63 | 1.48 | 0.04 | 0.05 | 0.11 |
| Baseline2 | 0.7 | 0.62 | 1.77 | 0.07 | 0.08 | 0.19 |
| Baseline2 refined | 0.43 | 0.29 | 0.92 | 0.07 | 0.08 | 0.19 |
| Template method | **0.22** | **0.18** | **0.55** | **0.03** | 0.08 | **0.05** |
| | | | Apricot Orchard, Summer | | | |
| Baseline1 | 0.54 | 0.55 | 1.6 | 0.05 | **0.05** | 0.16 |
| Baseline2 | 0.76 | 0.76 | 2.09 | 0.07 | 0.07 | 0.2 |
| Baseline2 refined | 0.43 | 0.35 | 0.94 | 0.07 | 0.07 | 0.2 |
| Template method | **0.26** | **0.22** | **0.64** | **0.03** | 0.07 | **0.06** |
| | | | Apricot Orchard, Winter | | | |
| Baseline1 | 0.92 | 1.1 | 2.85 | 0.07 | 0.09 | 0.21 |
| Baseline2 | 0.89 | 0.95 | 2.49 | 0.14 | 0.16 | 0.5 |
| Baseline2 refined | 0.49 | 0.64 | 1.23 | 0.14 | 0.16 | 0.5 |
| Template method | **0.22** | **0.17** | **0.53** | **0.02** | **0.05** | **0.04** |

† MAE: Mean Absolute Error;    SD: Standard Deviation;    95%: 95 Percentile;

*Table 6 Overall localization error for the all the localization methods in the vineyard and apricot in spring, summer, and winter.*

The improvement when using the template-based method over the baseline methods is more significant for the apricot orchard experiments than the vineyard experiments. The reason is that the grapevines (Figure 12 a) are trellised and form a "planar" surface, while the geometries of the apricot trees' canopies (Figure 12b, c, d) are much more irregular. Both baseline methods rely on line or parallel line fitting, which can be less accurate when the plant canopies do not form a wall-type planar surface. The template-based method exploits the entire point-cloud dataset for localization and is much less affected by irregular canopy shapes.

### 5.9.2 Accuracy comparison for large heading deviation

The localization results for large heading measurements are shown in Table 6. The large heading measurement only constitute 8.1% of all the measurement data in our experiment, however, the localization robustness at these cases is very important for robot navigation failure recovery.

| | Y error (meter) | | | Yaw error (rad) | | |
|---|---|---|---|---|---|---|
| **Method** | MAE† | SD† | 95%† | MAE | SD | 95% |
| | Vineyard | | | | | |
| Baseline1 | 1.06 | 1.19 | 2.28 | 0.08 | **0.06** | 0.20 |
| Baseline2 | 0.65 | 0.63 | 2.03 | 0.17 | 0.18 | 0.53 |
| Baseline2 refined | 0.80 | 0.42 | 1.54 | 0.17 | 0.18 | 0.53 |
| Template method | **0.16** | **0.17** | **0.41** | **0.03** | 0.09 | **0.06** |
| | Apricot Spring | | | | | |
| Baseline1 | 1.25 | 1.16 | 3.29 | 0.07 | 0.09 | 0.18 |
| Baseline2 | 0.98 | 0.94 | 2.54 | 0.13 | 0.13 | 0.44 |
| Baseline2 refined | 0.48 | 0.40 | 1.08 | 0.13 | 0.13 | 0.44 |
| Template method | **0.29** | **0.17** | **0.60** | **0.02** | **0.01** | **0.05** |
| | Apricot Summer | | | | | |
| Baseline1 | 0.95 | 0.79 | 2.67 | 0.08 | 0.07 | 0.2 |
| Baseline2 | 1.15 | 1.07 | 3.38 | 0.11 | 0.11 | 0.33 |
| Baseline2 refined | 0.52 | 0.54 | 1.2 | 0.11 | 0.11 | 0.33 |
| Template method | **0.24** | **0.18** | **0.56** | **0.02** | **0.02** | **0.05** |
| | Apricot Winter | | | | | |
| Baseline1 | 1.78 | 1.55 | 4.9 | 0.12 | **0.08** | 0.24 |
| Baseline2 | 0.95 | 1.34 | 2.99 | 0.23 | 0.19 | 0.53 |
| Baseline2 refined | 0.59 | 1.09 | 1.63 | 0.23 | 0.19 | 0.53 |
| Template method | **0.22** | **0.16** | **0.53** | **0.03** | 0.11 | **0.04** |

† **MAE: Mean Absolute Error;**     **SD: Standard Deviation;**     **95%: 95 Percentile;**

*Table 7 Localization error for large heading measurements (robot heading > 0.3rad) for the all the localization methods in the vineyard and apricot in spring, summer, and winter.*

We can see that compared to the overall localization results, all methods' localization results are degraded. However, the template-based method's accuracy degraded minorly while the baseline methods' localization accuracy degraded significantly. These results proved that the template-based method is more robust at large heading cases.

### 5.9.3  Comparison of accumulated error distributions

The accumulated error distributions for all methods are shown in Figure 29, in vineyard experiments, and in Figure 30, apricot orchard spring experiments, respectively. Apricot orchard summer experiments are shown in Figure 31, and apricot orchard winter experiments are shown in Figure 32. In all the figures, the curves represent the percentage of localization errors that are lower than the corresponding error threshold parameter of the x-axis. One can see that the template-based method is better than the baseline methods in all experiments, for both lateral and heading localization, and the advantage is more significant in non-wall shaped canopies (apricot orchard).

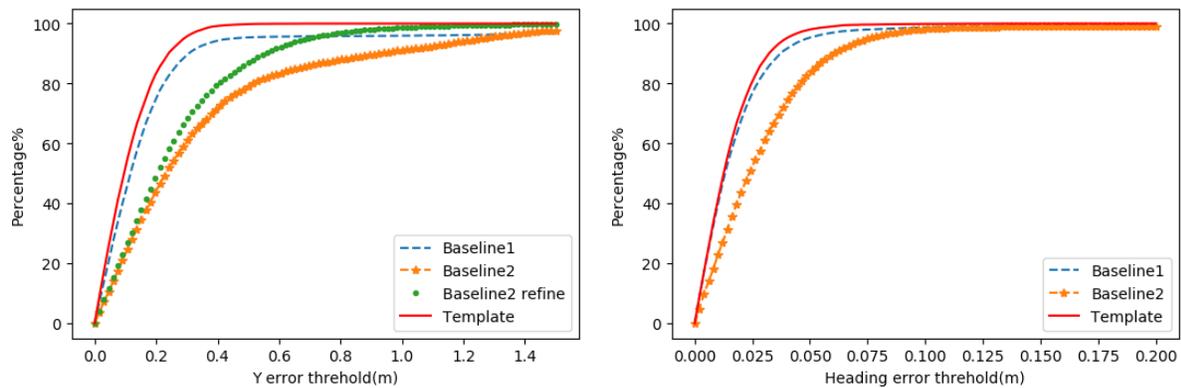

*Figure 29 Accumulated error distribution for all methods in vineyard experiments. Each curve shows the percentage of localization results that have error below the corresponding error threshold in x-axis. a) is for lateral error and b) is for heading error.*

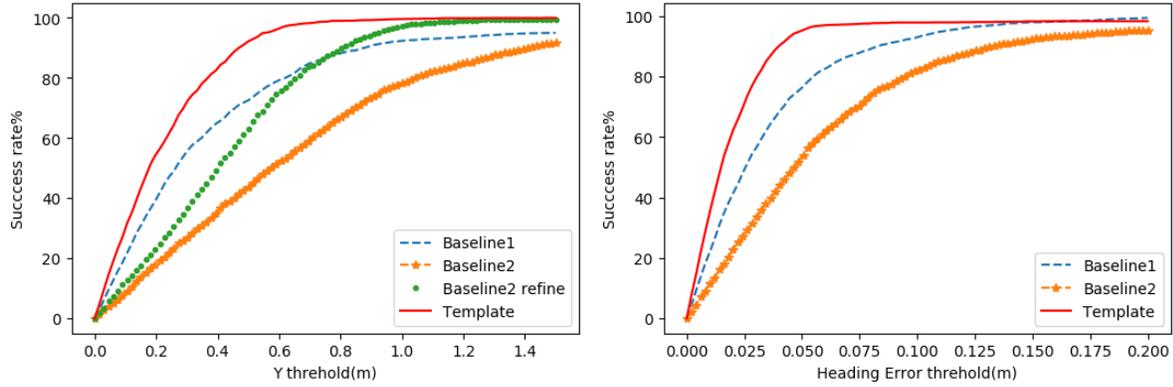

*Figure 30 Accumulated error distribution for all methods in apricot orchard spring experiments. Each curve shows the percentage of localization results that have error below the corresponding error threshold in x-axis. a) is for lateral error and b) is for heading error.*

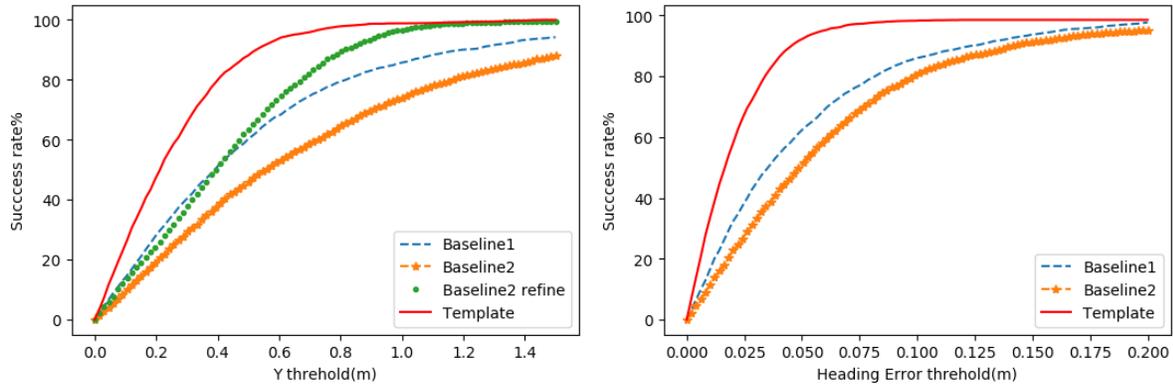

*Figure 31 Accumulated error distribution for all methods in apricot orchard summer experiments. Each curve shows the percentage of localization results that have error below the corresponding error threshold in x-axis. a) is for lateral error and b) is for heading error.*

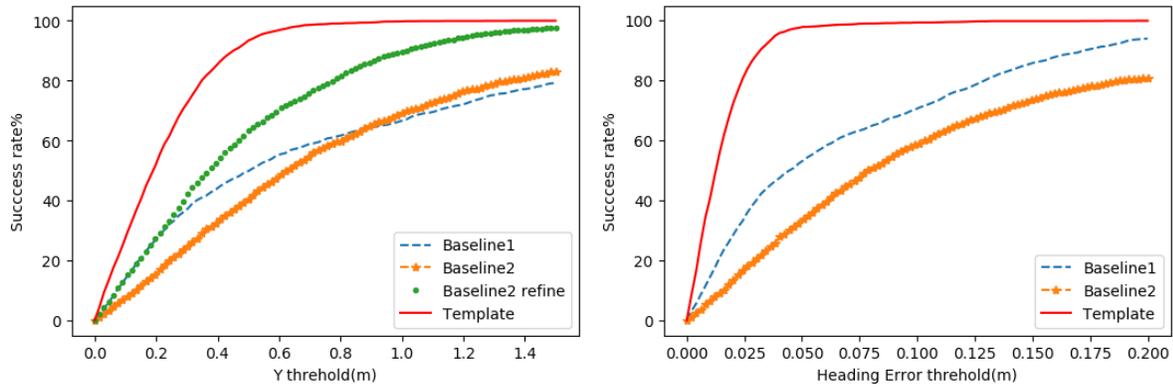

*Figure 32 Accumulated error distribution for all methods in apricot orchard winter experiments. Each curve shows the percentage of localization results that have error below the corresponding error threshold in x-axis. a) is for lateral error and b) is for heading error.*

## 5.9.4 Comparison of robustness against tree gaps in the rows

The template-based method was compared against the two baseline methods when trees were missing, resulting in gaps. The results are shown in Figure 33. As expected, the localization accuracy of all the methods decreased when the number of missing unit-trees increased. However, the template-based method performed always better than the baseline methods and was more robust to missing trees.

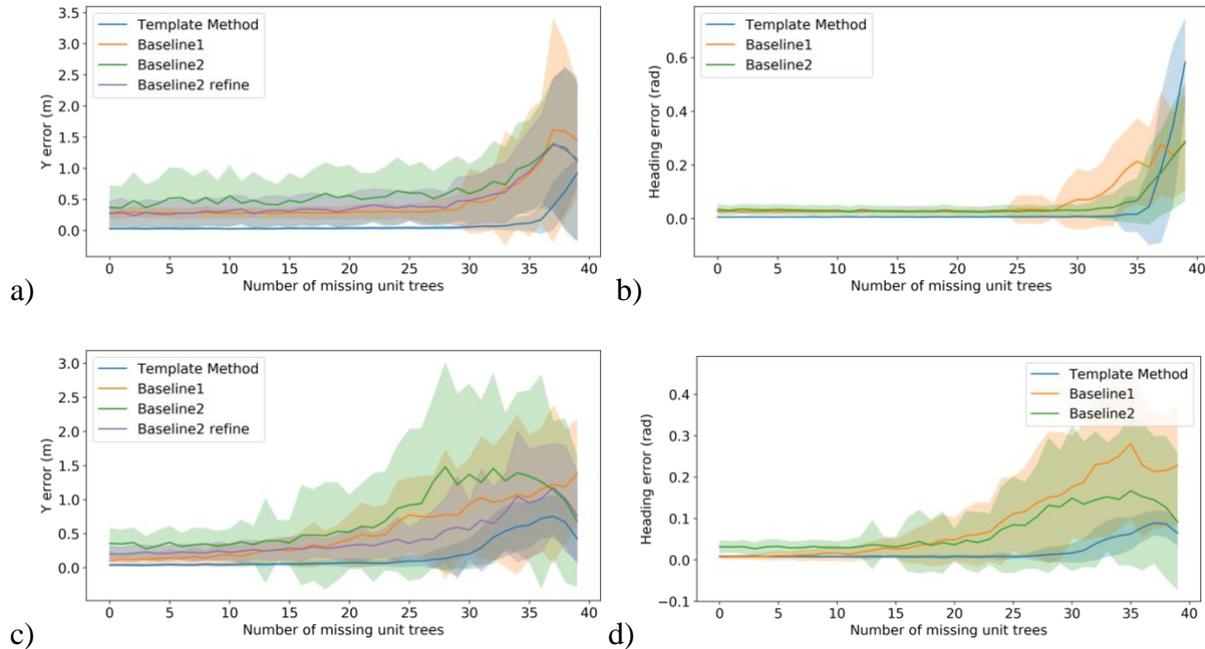

*Figure 33 Localization performance as a function of the number of random missing unit-trees at the sides of the row. a) Lateral error curves in the vineyard. b) Heading error curves in the vineyard. a) Lateral error curves in the apricot orchard. b) Heading error curves in the apricot orchard. The shaded regions indicate the error standard deviations.*

## 5.9.5 Comparison of localization near the row ends

The performance of the template-based method near a row's end (when the vehicle must exit the row) was compared against the baseline methods. The results are shown in Figure 34. As expected, the localization accuracy of all the methods decreased when the robot approached the end of a row, and all methods 'collapsed' after the distance to the row became too small (less than ~3m) to contain enough measurement points in the camera's field of view. However, the template-based method performed always better than the two baseline methods as the robot approached the row's end, and its performance was consistent up until 5 m to the end. The difference in performance was much more profound in the apricot orchard, most likely because the line-fitting methods did not have enough points to fit lines.

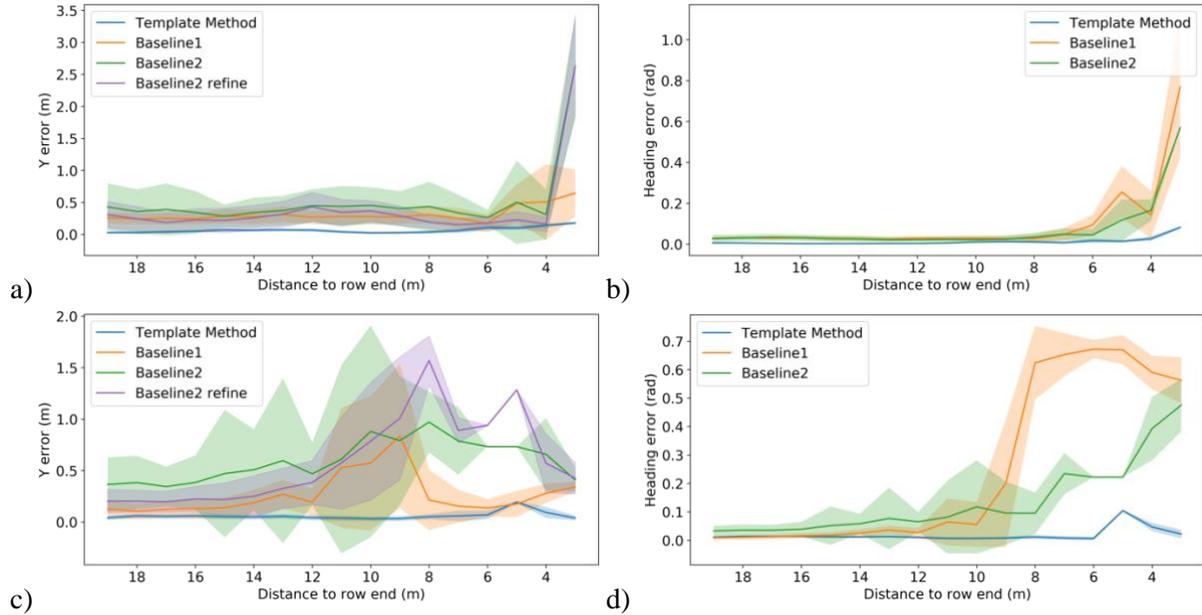

*Figure 34 The figures show the localization accuracy as a function of the robot's distance to the end of a row. a) Lateral error and b) heading error in the vineyard; c) lateral and d) heading error in the apricot orchard. The shaded regions indicate the error standard deviations.*

# 6 Conclusions and discussion

In this work, we proposed a generic 3D sensor-based method for robot localization with respect to orchard row centerlines. Instead of relying on assumptions about the presence - and detection ability - of features and their spatial distributions with respect to row centerlines, our method discovers orchard-specific structure in a data-driven way, encodes it as an orchard row-sensing template, and utilizes the full 3D measurement information to determine the vehicle pose. We also proposed a way to estimate the confidence of the template-based localization estimate. Experiments were performed in a vineyard, and in an apricot orchard in different seasons. The vineyard was an example of a fruit-wall trellised orchard, whereas the apricot orchard had trees that were spaced apart more, and had large non-planar canopies. Results showed that the method was quite accurate; lateral mean absolute error (MAE) was less than 3.6% of the row width, and heading MAE was less than 1.72°. Localization was also robust with respect to gaps in the tree rows, and the choice of row and number of measurements to build the template.

Our method assumes that the trees in the orchard are planted in a regular pattern (although it is robust to irregularities and gaps) and that the tree canopies are maintained reasonably well, so that the 3D sensor is not blocked and can access a 10-20 m segment of the tunnel-like canopy structure in front of it. These assumptions are realistic for most modern commercial orchards. However, further experiments would be needed to assess the method's performance under various orchard architectures and conditions.

One limitation of the proposed method – and actually of all in-row localization methods - is that the error grows significantly as the robot comes close to the end of the row. This happens because the set of sensed 3D points that belong to the current row becomes smaller, and the points beyond the end of the row can have arbitrary spatial distributions, which do not match the distribution of the points inside the row. This limitation could be overcome to some extent by using concurrently a front-looking and a rear-looking sensor – with their corresponding template localization threads running in parallel - and fusing their localization outputs based on the estimated confidence from each one.

The second limitation is that, in order to build the template, our method needs a set of initial measurements inside the row, with known ground truth sensor poses with respect to the centerline. In our vineyard experiments, GNSS signals were available - and used – for this purpose, whereas in the orchard, a colored rope was used as an easily detectable physical centerline. In real-world situations, accurate GNSS signals will not be available (if they are, the method is not needed) and stretching using a rope is impractical. One possibility is to setup a small number (e.g., 2 to 4) of permanent, easy-to-sense artificial landmarks in some location inside an orchard row and use them to compute the sensor's true pose with respect to the centerline and build the template. The computation could be automated, and the template could be built while driving a short distance (e.g., 5 m) with the landmarks in clear sight. Future work will investigate the possibility of using a single frame (see Table 4) to build a template and incrementally update the template as the robot travels in the orchard. Using a small set of landmarks is a practical and general approach, because it does not rely on any assumptions about the appearance of the orchard rows or the robot. The approach is also adaptive, because it enables automatic adaption/update of row-templates when tree geometries/appearance change because of events such as season changes, pruning or thinning. One can think of this approach as creating a '*home position*' for the mobile robot, analogous to the fixed home position of a manipulator arm, where all joint angles are known.

Overall, the experimental results indicate that the proposed localization method is accurate and robust, and by re-building the template, the method can adapt to different, well managed modern straight row orchards and to dynamic changes in orchard appearance. Thus, the proposed method presents a generic approach to localization inside orchard rows, and in principle, inside any agricultural and non-agricultural structures that exhibit characteristics #1 and #2; such environments include row-crops, greenhouse tunnels, corridors, aisles, etc.

## Acknowledgments
This work was funded partially by NIFA Grant 2020-67021-30759, under the National Robotics Initiative, and by NIFA Hatch/Multi-State Grant 1001069. We thank our research and development engineer Dennis Sadowski for building the experiment robot and thank Chen Peng and Ben Gatten for their help in the experiments.